\documentclass[10pt,twocolumn,letterpaper]{article}

\usepackage{amsmath}
\usepackage[hang,flushmargin]{footmisc} 
\usepackage{cvpr}              

\newcommand{\szh}[1]{{\color{black}#1}}

\usepackage{gensymb}
\usepackage[colorlinks=true, urlcolor=blue, linkcolor=red]{hyperref}
\usepackage{amsmath}

\DeclareMathOperator*{\argmin}{arg\,min}

\usepackage{siunitx}
\usepackage{adjustbox}
\usepackage{multirow}
\usepackage{colortbl}
\usepackage{xcolor}
\definecolor{mygray}{gray}{0.8}

\usepackage{pgffor}
\newcommand{\widthscale}{0.138}

\def\figvspace{{\vspace{-3mm}}}

\usepackage{wrapfig}

\definecolor{cvprblue}{rgb}{0.21,0.49,0.74}

\def\paperID{1355} 
\def\confName{CVPR}
\def\confYear{2025}

\title{IMFine: 3D \underline{I}npainting via Geometry-guided \underline{M}ulti-view Re\underline{fine}ment}

\author{
Zhihao Shi$^{*1}$~~~~~~~~~~Dong Huo$^{*\circ2}$~~~~~~~~~~Yuhongze Zhou$^{\dagger1}$~~~~~~~~~~
Kejia Yin$^{\dagger1}$\\
Yan Min$^{\circ3}$~~~~~~~~~~Juwei Lu$^{1}$~~~~~~~~~~Xinxin Zuo$^{4}$\\
$^1$Huawei Canada Research Institute~~~~~~~~~~$^\text{2}$University of Alberta\\$^\text{3}$McMaster University~~~~~~~~~~$^\text{4}$Concordia University\\
{\tt\small \
zhihaoshi2022@gmail.com, dhuo@ualberta.ca, yuhongze.zhou@mail.mcgill.com}\\
{\tt\small \
kejia.yin@mail.utoronto.ca, miny13@mcmaster.ca, juwei.lu@huawei.com, xinxin.zuo@concordia.ca}
}

\begin{document}

\twocolumn[{
 \maketitle
\centerline{
 \includegraphics[width=\linewidth]{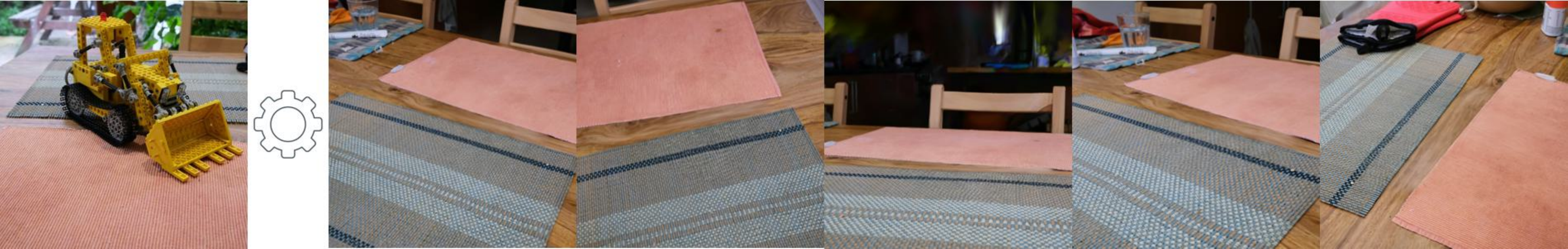}
}
\centerline{
 \includegraphics[width=\linewidth]{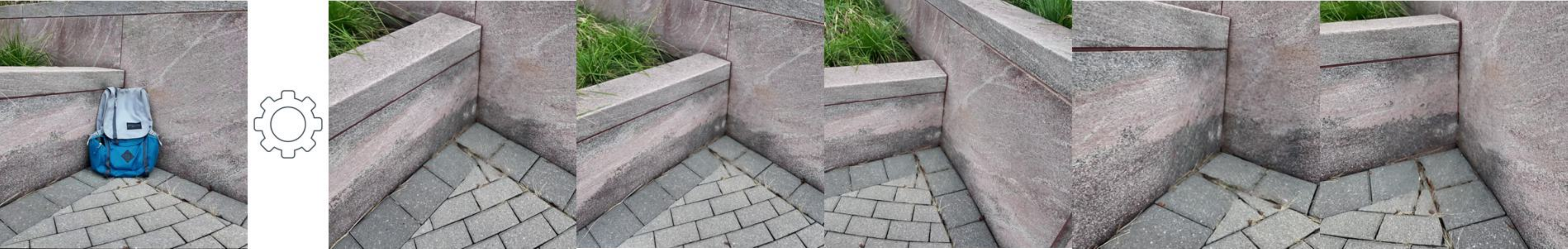}
}
\captionof{figure}{
We present a novel 3D inpainting method capable of removing target objects from the 3D scenes and seamlessly inpainting the occluded regions across various scenarios. The proposed method ensures the generation of plausible geometry and consistent textures. \textit{It is highly recommended to see video results in our \href{https://xinxinzuo2353.github.io/imfine}{\textbf{project page}}}.}
\label{fig:teaser}
\vspace{4mm}}]

\def\thefootnote{}
\footnotetext{$*$ and $\dagger$ denote equal contribution.}
\footnotetext{$\circ$ Work done during an internship at Huawei Canada Research Institute}

\begin{abstract}
Current 3D inpainting and object removal methods are largely limited to front-facing scenes, facing substantial challenges when applied to diverse, “unconstrained" scenes where the camera orientation and trajectory are unrestricted.
To bridge this gap, we introduce a novel approach that produces inpainted 3D scenes with consistent visual quality and coherent underlying geometry across both front-facing and unconstrained scenes. 
Specifically, we propose a robust 3D inpainting pipeline that incorporates geometric priors and a multi-view refinement network trained via test-time adaptation, building on a pre-trained image inpainting model.
Additionally, we develop a novel inpainting mask detection technique to derive targeted inpainting masks from object masks, boosting the performance in handling unconstrained scenes. 
To validate the efficacy of our approach, we create a challenging and diverse benchmark that spans a wide range of scenes. Comprehensive experiments demonstrate that our proposed method substantially outperforms existing state-of-the-art approaches.

\end{abstract}  
\section{Introduction}
\label{sec:intro}
3D editing is an active research area with significant potential for various applications, including video games, the Metaverse, virtual reality, and augmented reality. For example, after a 3D scene is created or reconstructed, a user might want to remove certain objects from the scene. This paper mainly focuses on the challenging task of inpainting in 3D scenes to fill the occluded regions to enable seamless object removal.

Compared to image inpainting ~\cite{suvorov2022resolution, rombach2022high}, 3D inpainting models lag significantly behind due to the lack of large-scale 3D datasets and efficient 3D data structures. Therefore, existing 3D inpainting methods mainly focus on lifting 2D inpainting priors to 3D.
Although the current diffusion-based image inpainting models can generate images with vivid textures and high-fidelity details, they are trained on single images without considering any 3D consistency. It poses great challenges in maintaining view consistency and preserving 3D coherence when lifting those 2D priors to 3D. 

To address the inconsistency issue, there are approaches that implicitly lift the 2D models into 3D via score distillation sampling~(SDS) loss ~\cite{mirzaei2024reffusion, chen2024mvip}, which was proposed initially in text-to-3D generation~\cite{poole2022dreamfusion}, to harness the inherent coherence of 2D diffusion models under the same prompt, complemented by the iterative updating scheme during optimization to ensure view consistency. 
However, they are constrained by the inherent limitation of SDS loss, which results in over-saturation and over-smoothing in the optimized images, even with different variants of SDS losses.
Alternatively, explicit methods ~\cite{mirzaei2023spin, chen2024gaussianeditor, mirzaei2023reference, wang2024innerf360, cao2024mvinpainter} first generate one or few images with the image inpainting models and try to preserve the 3D consistency by enforcing geometry constraints or minimizing perceptual losses between rendered images and the inpainted 2D images. Although they have demonstrated good performance, especially for front-facing scenes with very limited view changes, the inpainted 3D scenes have artifacts such as floaters and blurry textures in unconstrained scenarios with complex geometry and large view variation. The reasons are that first, 3D geometry is rather difficult to get perfectly restored;
second, perceptual loss can only alleviate the inconsistency by minimizing overall similarity. Inconsistency still exists between those inpainted images, leading to blurry textures for the inpainted 3D scenes. 
More recently, ~\cite{cao2024mvinpainter} proposed to train a multi-view inpainting network to generate consistent multi-view inpainted images. However, due to the limited diversity of current multi-view datasets, they cannot generalize to unconstrained 3D scenes~(e.g., 360\degree cases).


To enhance the ability to generalize to various scenes for 3D inpainting, we utilize geometry priors and develop a multi-view refinement network fine-tuned at test time to produce consistent images in a coarse-to-fine manner.
Specifically, we start with a single reference view inpainted with image inpainting models, followed by coarse geometry restoration and a warping operation to warp the inpainted content from the reference view to others.
However, due to the imperfect reconstructed geometry and view-dependent effects, the warped images tend to have artifacts, such as distorted textures and mismatches between the warped and neighboring contents.
To refine the warped contents, we present a multi-view refinement model to generate consistent images with high fidelity, taking the warped images as a condition. The key idea is that instead of training a model with extensive datasets, we take the pre-trained image inpainting model, add space-time attention layers, and fine-tune them with the current test scenes. We take the originally captured multi-view images and implement various augmentations to synthesize paired data to fine-tune the multi-view refinement model.

Furthermore, previous works usually take the entire object masks as inpainted regions. However, some regions inside the object masks might have already been exposed in other views, especially for 3D object removal and inpainting in 360\degree scenes (see Fig.~\ref{fig:mask and detection}). Therefore, we believe it is critical to differentiate between object masks and inpainting masks, fully exploiting the available information of the reconstructed 3D scenes. To this end, a novel inpainting mask detection approach is proposed to extract masks accurately and exclusively for real occluded regions.

Finally, considering that the existing benchmarks mostly cover front-facing scenes or planar surfaces, we introduce a new dataset encompassing 20 diverse scenes, from indoor to outdoor environments, covering large view variations and complex interfacial surfaces, to better evaluate and advance 3D inpainting methods.

\noindent In summary, our contributions are as follows:
\begin{enumerate}
    \item A multi-view consistent inpainting pipeline is proposed. We leverage geometric \szh{priors} and a multi-view refinement model that is fine-tuned from image inpainting models via test-time adaptation to \szh{generate consistent multi-view images}. 
    \item We further develop an inpainting mask detection technique that accurately distinguishes occluded regions, enabling robust processing of complex scenes, such as 360\degree environments.
    \item The proposed inpainting method has achieved state-of-the-art performance across various benchmarks, particularly excelling in challenging, unconstrained scenarios.
    \item We contribute a new and challenging 3D scene inpainting benchmark, offering diverse scenes to facilitate broader evaluation and advancement in 3D inpainting research.
\end{enumerate}

\section{Related Work}
\label{sec:relatedwork}

\subsection{3D Inpainting}
Existing 3D inpainting methods \cite{chen2024gaussianeditor, wang2024innerf360, mirzaei2023spin, mirzaei2023reference, wang2024gscream, chen2024mvip, mirzaei2024reffusion} are implemented by lifting 2D inpainting priors to 3D with various strategies designed to ensure 3D consistency.
One direction, inspired by score distillation-based approaches \cite{poole2022dreamfusion} in text-to-3D generation, formulates 3D inpainting as a continuous distillation process \cite{mirzaei2024reffusion, chen2024mvip}. While these approaches achieve reasonably consistent inpainting results, they face inherent limitations of score distillation, such as producing over-saturated or overly smooth outputs.
Another direction explicitly enforces 3D consistency by independently inpainting one or multiple 2D views and then consolidating multi-view inconsistencies in 3D. Reference-guided NeRF \cite{mirzaei2023reference} addresses this challenge by using a single view to guide geometry inpainting to avoid inconsistencies. However, it struggles with accurate geometry capture in complex 360\degree scenes using just a single view.
SPINeRF \cite{mirzaei2023spin}, and InNeRF360 \cite{wang2024innerf360} utilize multiple inpainted images to guide 3D inpainting and incorporate perceptual loss to reduce inconsistencies resulting from independently inpainted images. Nevertheless, while perceptual loss can maintain overall consistency, it performs poorly on finer details, often leading to blurry 3D results.
Our method lies in the latter direction. It addresses these issues through a 
geometry-guided refinement network that generates a set of multi-view consistent images, reconstructing clearer and consistent 3D inpainted scenes.

\subsection{Video and Multi-View Diffusion}
The diffusion model for text-to-image (T2I) generation \cite{ho2020denoising, ho2022classifier, song2020denoising, rombach2022high} has achieved significant success in recent years, known for its ability to produce high-quality, diverse, and controllable outputs. Building on this, extensive studies have extended image diffusion models to the video generation domain \cite{ho2022imagen, ho2022video, singer2022make, zhou2022magicvideo} by incorporating temporal layers within each transformer block to perform space-time attention, which captures correlations across frames and shows promising results in generating realistic videos.

Inspired by these advances in video generation, researchers have begun exploring the adaptation of T2I models into multi-view diffusion models for implicit 3D reasoning \cite{shi2023mvdream, shi2023zero123++, wang2023imagedream, liu2023syncdreamer}. For example, methods such as MVDream \cite{shi2023mvdream} and Zero123++ \cite{shi2023zero123++} leverage this idea and derive multi-view diffusion models, capable of generating consistent multiple views simultaneously for 3D generation. However, the full space-time attention mechanism used in these methods can lead to prohibitive memory and computational costs, making it impractical for handling large numbers of views.
To address this, recent work \cite{wu2023tune} introduces a sparse space-time attention mechanism, which reduces memory requirements by attending only to the initial and previous frames. However, the sparse attention approach assumes input frames are from a continuous video sequence and remains under-explored in sparse multi-view scenarios. 

\subsection{Diffusion Model Fine-tuning}
Adapting large pre-trained diffusion models has proven to be an effective and efficient approach for completing user-specified tasks, such as synthesizing specific subjects in new contexts \cite{ruiz2023dreambooth, chen2024anydoor} and inpainting images with instance matching reference images \cite{tang2024realfill, yang2023paint}. Adaptation methods include injecting extra layers \cite{mou2024t2i, ye2023ip, zhang2023adding, yang2023paint}, searching appropriate unique prompt \cite{ruiz2023dreambooth}, updating partial or full model weights \cite{mirzaei2024reffusion, lin2025taming, tang2024realfill}, or combinations of these techniques.
Similarly, in this paper, instead of relying on extensive multi-view datasets, we propose adapting the pre-trained image diffusion model into a multi-view refinement model for 3D inpainting purposes.


\section{Preliminary}
Gaussian splatting~(GS) methods \cite{kerbl20233d, huang20242d} represent a scene as a collection of multivariate Gaussian primitives:
    $\mathcal{G} = \{
        (\mu_m, \Sigma_m, \Gamma_m, \text{SH}_m)
    \}_{m = 1}^M,$
where $\mu_m$, $\Sigma_m$, $\Gamma_m$ and $\text{SH}_m$ denote the mean vector, covariance matrix, opacity, and spherical harmonic coefficients of a multivariate Gaussian, respectively.
View synthesis is performed by standard point-based $\alpha'$-rendering, where a single pixel color is computed by sorting $\mathcal{G}$~(or their 2D projection) in their depth order and alpha-blending each contributed primitive:
\begin{equation}
    C = \sum_{m} c_m\alpha_m\prod_{j = 1}^{m - 1} (1 - \alpha_j),
\label{eq:alpha_rendering}
\end{equation}
where $C$ is the output pixel color, $c_m$ is the pre-computed color of the $m$-th primitive, and $\alpha_m$ is a calculated opacity based on distance from the Gaussian center to the interaction point. 
Due to the excellent performance in geometry reconstruction, in this work, we choose 2D-GS \cite{huang20242d} as the 3D representation.

\section{Method}
\begin{figure*}[t]
    \centering
    \begin{subfigure}{1\textwidth}
        \centering
        \includegraphics[width=\linewidth]{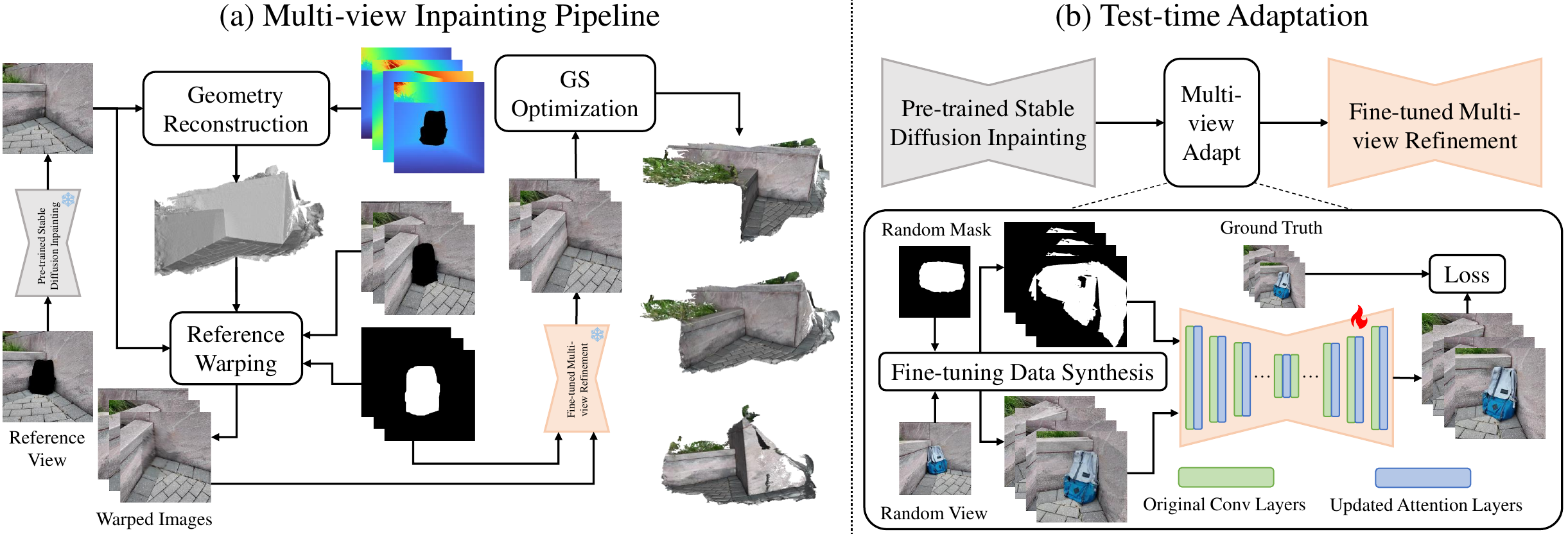} 
        \label{fig:image1}
    \end{subfigure}
    \vspace{-8mm}
    \caption{
    \textbf{Overview of our proposed approach.} (a) shows the proposed 3D inpainting pipeline. Specifically, a reference view is first inpainted, which will be used to reconstruct complete geometry, along with rendered multi-view depth maps. The inpainted reference image is then warped to other views, which are taken as input conditions for the multi-view refinement network to generate a set of view-consistent images. Afterward, we fine-tune the 3D scene to get the inpainted scene. (b) illustrates the detailed structure and training of the multi-view refinement network. Basically, we take the pre-trained image inpainting model, incorporate additional space-time attention layers, and exploit the test-time adaptation to fine-tune it into a multi-view refinement model. To generate a fine-tuning dataset, we synthesize warped images under the original captured scene, which has the ground-truth captured images. In detail, reference views are selected randomly, and we set random masks over the images, apply geometry jittering, and warp the reference images in other views.
    }
    \vspace{-5mm}
    \label{fig:pipeline}
\end{figure*}

Given a GS scene $\mathcal{G}$, along with corresponding training images $\{I_n\}_{n=1}^N$, associated camera poses $\{\Pi_n\}_{n=1}^N$, and a set of object masks $\{M_n\}_{n=1}^N$ identifying objects to be removed, our goal is to generate an inpainted scene $\mathcal{G}^{\ast}$ where the target objects are removed and occluded regions are filled seamlessly with the background. The entire pipeline contains two stages, namely object removal and 3D scene inpainting. 
\szh{We first remove object primitives (Sec.~\ref{sec:remove}), exposing never-before-seen~(NBS) regions which are the areas that truly need inpainting. 
We then either use the object masks as inpainting masks when they are identical or detect specific inpainting masks (Sec.~\ref{sec:additional}), ensuring that the inpainting mask exclusively covers the NBS regions in each view, denoted as $\{P_n\}_{n=1}^N$.
Given the pruned GS scene and inpainting masks, we propose a novel 3D inpainting pipeline to inpaint $\mathcal{G}$ as $\mathcal{G}^{\ast}$ (Sec.~\ref{sec:pipe}).}

\subsection{Object Removal}
\label{sec:remove}
To remove the objects from the reconstructed GS scene, we append an additional learnable parameter $\{p_m\}_{m=1}^M \in [0,1]$ to each Gaussian primitive, indicating whether it belongs to the object to be removed, which is initialized as $0.5$. To compute the value of $p_m$ for each Gaussian, we map the object masks back to the Gaussian primitives. Specifically, we render a set of object masks $\{\hat{M}\}_{n=1}^N$ using $\alpha'$ rendering (similar to color rendering) at $\{\Pi_n\}_{n = 1}^N$ but replacing $c_m$ with $p_m$ in Eq.~\ref{eq:alpha_rendering}, followed by $K_1$ ($K_1=1000$) steps of the optimization to update $p_m$ with respect to $\mathcal{L}_1$ loss between $\{\hat{M}\}_{n=1}^N$ and $\{M\}_{n=1}^N$. Notably, other parameters remain unchanged during this procedure. Finally, a pruning operation is performed to produce the pruned scene $\mathcal{G}_{\text{pruned}}$ by thresholding $p_m$ with a threshold $\tau$~($\tau=0.4$).


\subsection{Proposed 3D Inpainting Pipeline}
\label{sec:pipe}
Given the pruned GS scene $\mathcal{G}_{\text{pruned}}$ as well as inpainting masks $\{P_n\}_{n=1}^N$ indicating the regions to be inpainted for each view, we propose a novel 3D inpainting pipeline shown in Fig~\ref{fig:pipeline}. 
\szh{We first render color images $\{I_n^{'}\}_{n = 1}^N$ from $\mathcal{G}_{\text{pruned}}$ for each view and select a reference view with image $I_r^{'} \in \{I_n^{'}\}_{n = 1}^N$, which is then inpainted using a pre-trained inpainting model \cite{rombach2022high} to reference image $I_r^{\ast}$}. We conduct geometry reconstruction to fill out the hole in 3D, which will be utilized to warp the inpainted contents in the reference image into the remaining views. 
\szh{We employ a multi-view refinement model to further refine the warped images, ensuring high fidelity and consistency, and then update the 3D scene.}
The model is fine-tuned from the pre-trained image inpainting model via a test-time adaptation strategy. The geometry reconstruction and multi-view refinement model are described as follows. 

\subsubsection{Geometry Reconstruction}
Following \cite{huang20242d}, we render a set of depth maps $\{D_n\}_{n = 1}^N$ from the prune GS scene $\mathcal{G}_{\text{pruned}}$ and adopt a truncated signed distance function (TSDF)~\cite{curless1996volumetric} technique to reconstruct the 3D mesh. However, $\{D_n\}_{n = 1}^N$ only contributes to the reconstruction of the background region, as the depth values within their inpainting mask areas are invalid. 
To fill the masked region, we aim to generate an all-valid depth map $D_r^{\ast}$ of the reference view by filling the masked areas in $D_r$ based on information from $I_r^{\ast}$. The filled depth map should meet two key requirements: 1) it must be consistent with the reference image $I_r^{\ast}$; 2) it should seamlessly connect with the depth values outside the masked region.

We begin by exploiting a monocular depth estimator \cite{depthanything} to estimate depth $\Bar{D_r}$ from $I_r^{\ast}$, followed by a linear transformation to convert $\Bar{D_r}$ to scene scale $\Tilde{D_r}$: $\Tilde{D_r} = a * \Bar{D_r} + b$, where the parameters $a,b$ are computed in the neighborhood of $P_r$ to minimize the difference between $\Bar{D_r}$ and $D_r$ in advance, following \cite{mirzaei2023reference}. We then develop a filling scheme according to $\Tilde{D_r}$. Formally, the filled values of $D_r^{\ast}$ are computed as follows:
\begin{align}
    D_r^{\ast}(x) = &\argmin_{D_r^{\ast}(x)} \sum_{x \in P_r}||\triangledown D_r^{\ast}(x) - \triangledown \Tilde{D_r}(x)||^2 + \lambda \triangledown D_r^{\ast}(x)\\\notag
    &\text{s.t.} \hspace{1mm} D_r^{\ast}(x ) = D_r(x),\text{where } P_r(x) = 0,
\end{align}
where $\triangledown$ denotes the Laplacian caculation and $x$ represents the pixel coordinate. Intuitively, the first term encourages $D_r^{\ast}$ to maintain a similar changing trend as $\Tilde{D_r}$, while the second term regularizes $D_r^{\ast}$ to be smooth and consistent with the depths outside the mask. 
As such, $D_r^{\ast}$ can be used for the reconstruction of the NBS region.

\subsubsection{Test-Time Adaptation for Multi-view Refinement}
\begin{figure}[t]
    \centering
\begin{subfigure}{0.15\textwidth}
        \centering
        \includegraphics[width=\linewidth]{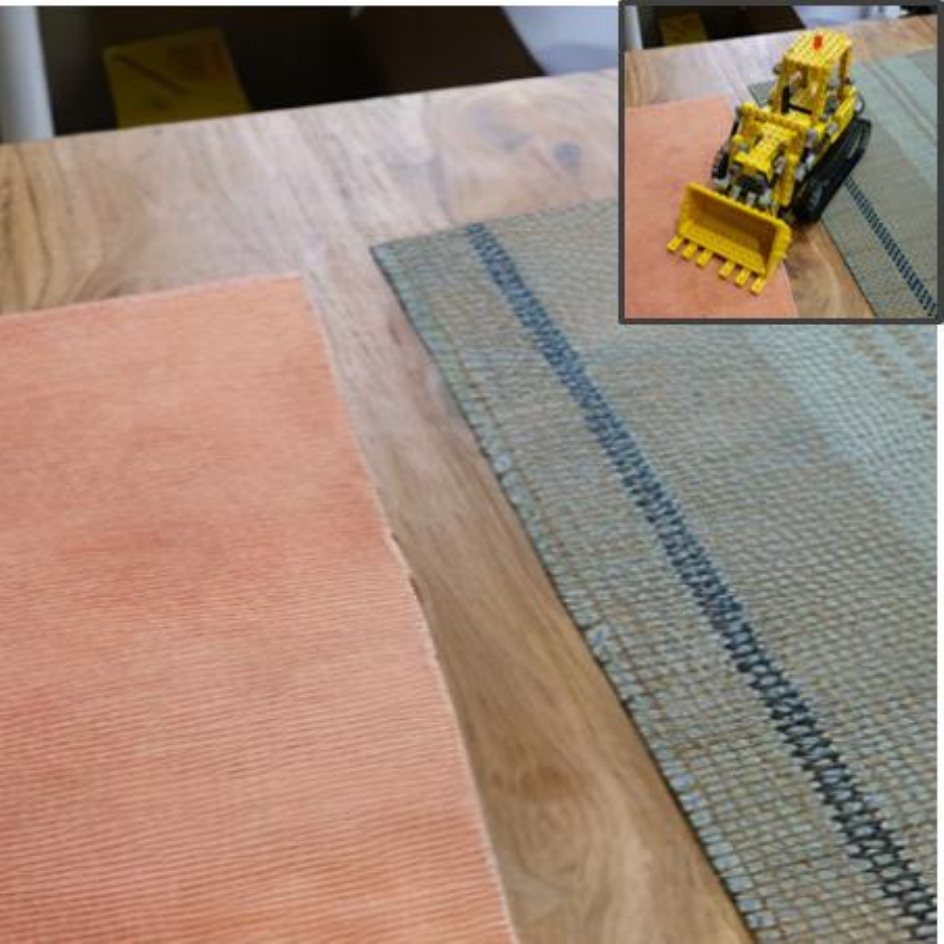} 
        \caption{Reference Image}
    \end{subfigure}
    \begin{subfigure}{0.15\textwidth}
        \centering
        \includegraphics[width=\linewidth]{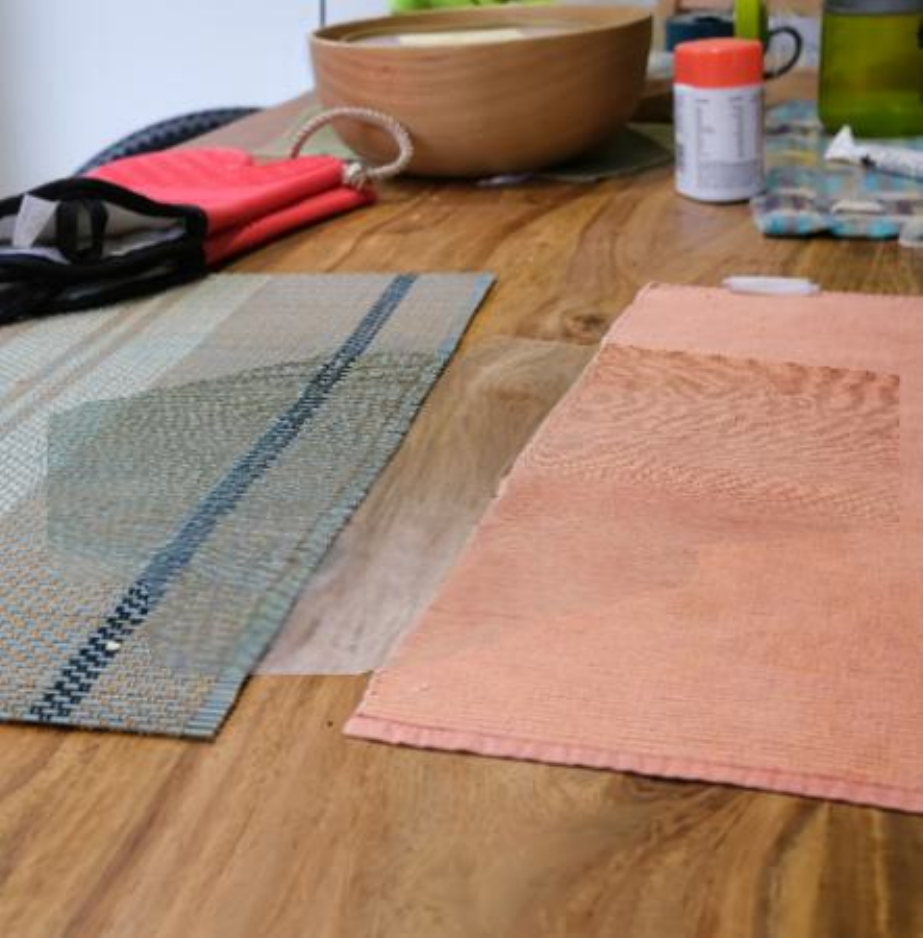} 
        \caption{Warped Image}
    \end{subfigure}
    \begin{subfigure}{0.15\textwidth}
        \centering
        \includegraphics[width=\linewidth]{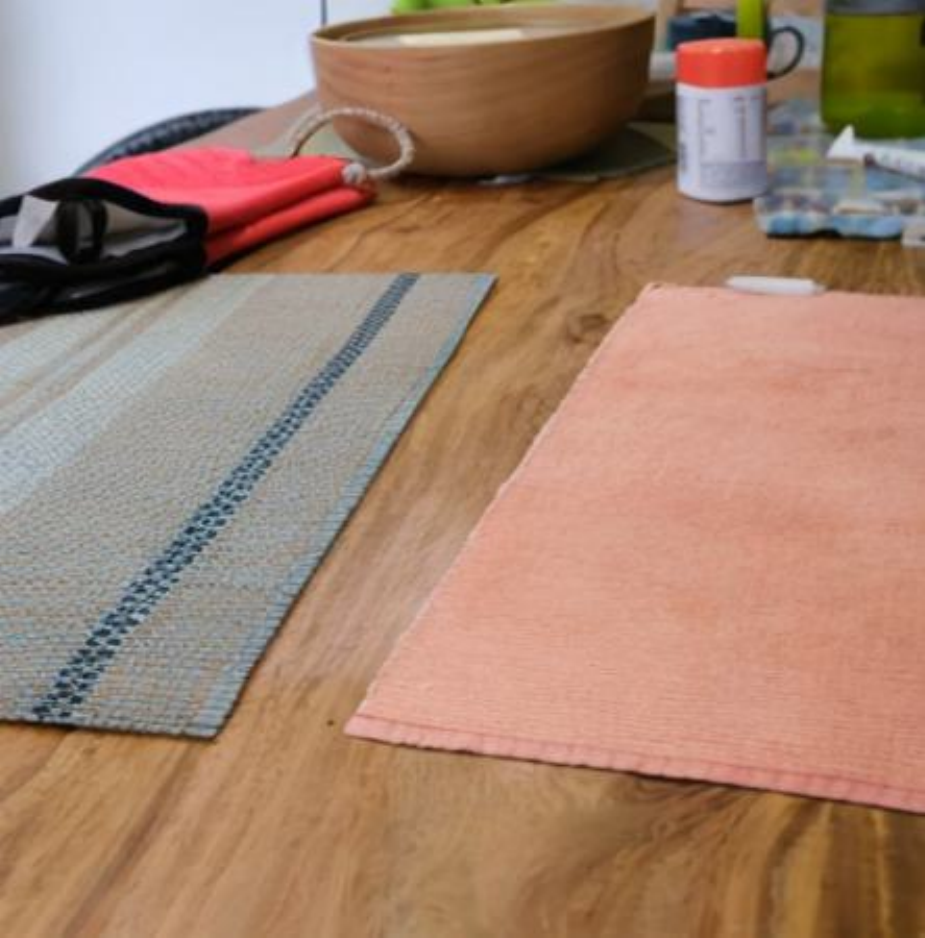} 
        \caption{Refined Image}
    \end{subfigure}
    \figvspace
    \caption{\textbf{Illustration of the importance of multi-view refinement.} The multi-view refinement model is proposed to tackle the warping artifacts with the example shown in (b), generating high-fidelity multi-view consistent images with one sampled view shown in (c).}
    \label{fig:warp_refine}
    \vspace{-5mm}
\end{figure}

We warp the inpainted contents from the reference view $I_r$ to other views with the above restored geometry. Due to the imperfect geometry and the neglect of view-dependent effects, those warped contents contain artifacts, such as texture distortion and inconsistent appearance with the neighboring regions, especially when there are large view angle changes with respect to the reference view as shown in Fig~\ref{fig:warp_refine}. On the other hand, these warped contents have coarse-level consistency across views. To this end, we propose a multi-view refinement model that produces high-fidelity, consistent images conditioning on the warped images. 

\noindent \textbf{Network:} Our model is built upon a pre-trained diffusion-based image inpainting model \cite{rombach2022high}, which takes a masked image and its associated mask as the input, corresponding to the warped image and inpainting mask in our scenarios, respectively. 
To ensure multi-view consistency, we replace the self-attention layer in each transformer block with space-time attention to explore correlations across multiple views. 
Considering the quadratic complexity of global attention and the large number of warped images, space-time attention is implemented as the sparse space-time attention \cite{ho2022video} that only attends to neighboring images. The attention layer is also allowed to attend to the reference image for alignment.

\noindent \textbf{Training:} Instead of relying on large datasets and extensive training, we exploit the test-time adaptation strategy for each scene to train the multi-view refinement network. Paired training sets are created for model adaptation. Specifically, given the original images $\{I_n\}_{n=1}^N$, at each training step, $L + 1$ images are randomly selected as ground truth. To simulate realistic warping scenarios, we randomly choose one image as the reference, generate a random mask, and warp both the reference image and mask to other views with geometry guidance from $\mathcal{G}$.
In addition, to encourage cross-view learning as in \cite{cao2024mvinpainter}, we also synthesize another set of paired images by applying image augmentations independently on each captured image, i.e., irregularly shaped masks that are randomly generated on each image. Image-based augmentations—such as elastic transformations and color jittering, are then applied to the masked regions to simulate warping artifacts.
We train the multi-view refinement model with those paired images, employing a simplified variational bound objective \cite{ho2020denoising}.

\noindent \textbf{Inference:} At inference, similar to ~\cite{weber2024nerfiller}, we randomly shuffle the order of the $L$ warped images at each denosing step to break the sequential constraint of sparse attention, further enhancing cross-view consistency.


\subsubsection{Fine-tune GS Scene}
\label{sec:finetune}
With the generated multi-view consistent images, we finetune $\mathcal{G}_{\text{pruned}}$ for $K_2 (K_2=7000)$ steps with an $L_1$ loss and a SSIM loss as straightforward pixel-wise supervision.
It is worth noting that other robust losses (e.g., perceptual loss) are functionally orthogonal to our approach and can be used to improve it further. 

\begin{figure}[t]
    \centering    
    \begin{subfigure}{0.23\textwidth}
        \centering
        \includegraphics[width=\linewidth]{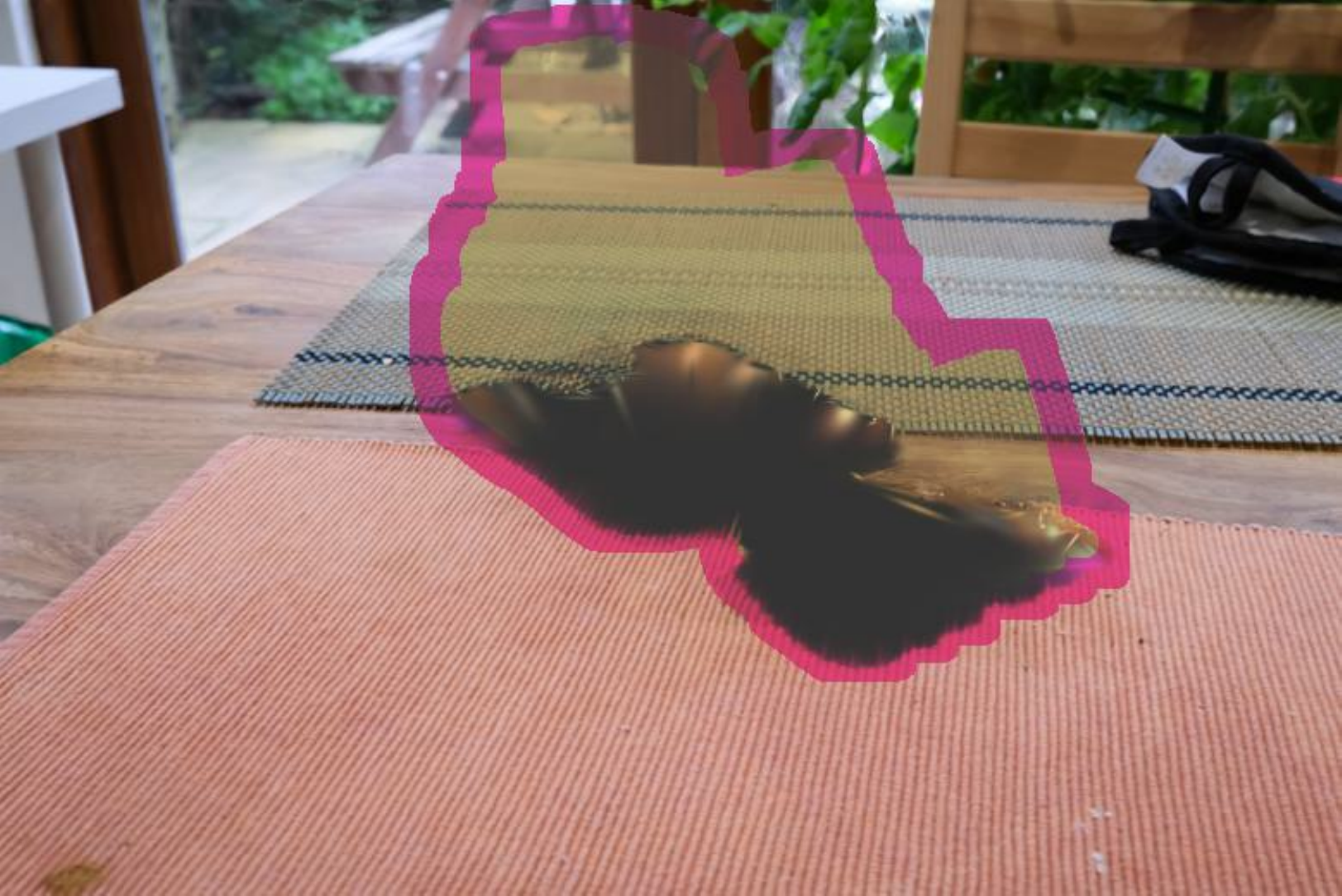} 
        \caption{Object mask and dilated mask}
    \end{subfigure}
    \begin{subfigure}{0.23\textwidth}
        \centering
        \includegraphics[width=\linewidth]{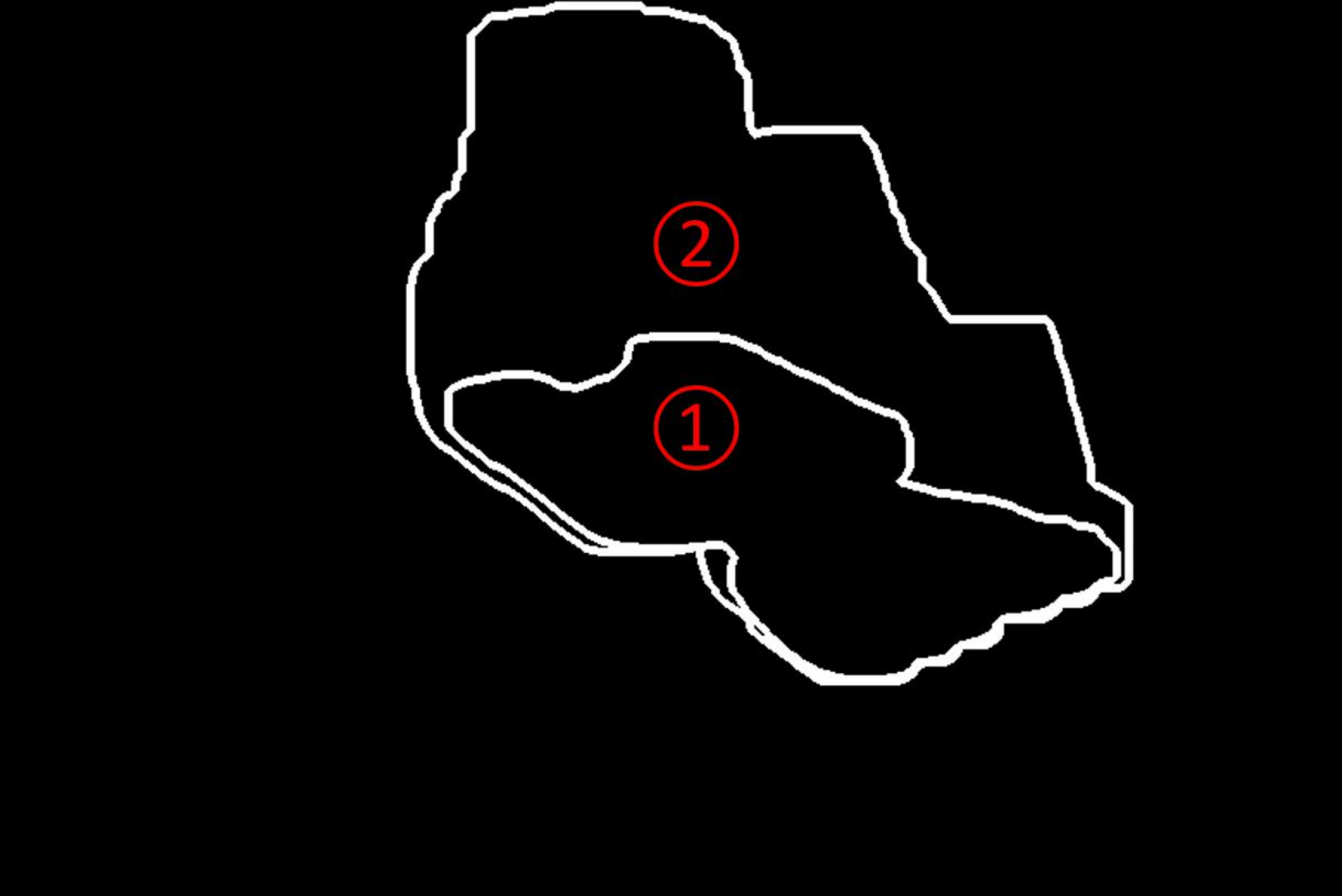} 
        \caption{Object Mask Seperation}
    \end{subfigure}
    \begin{subfigure}{0.23\textwidth}
        \centering
        \includegraphics[width=\linewidth]{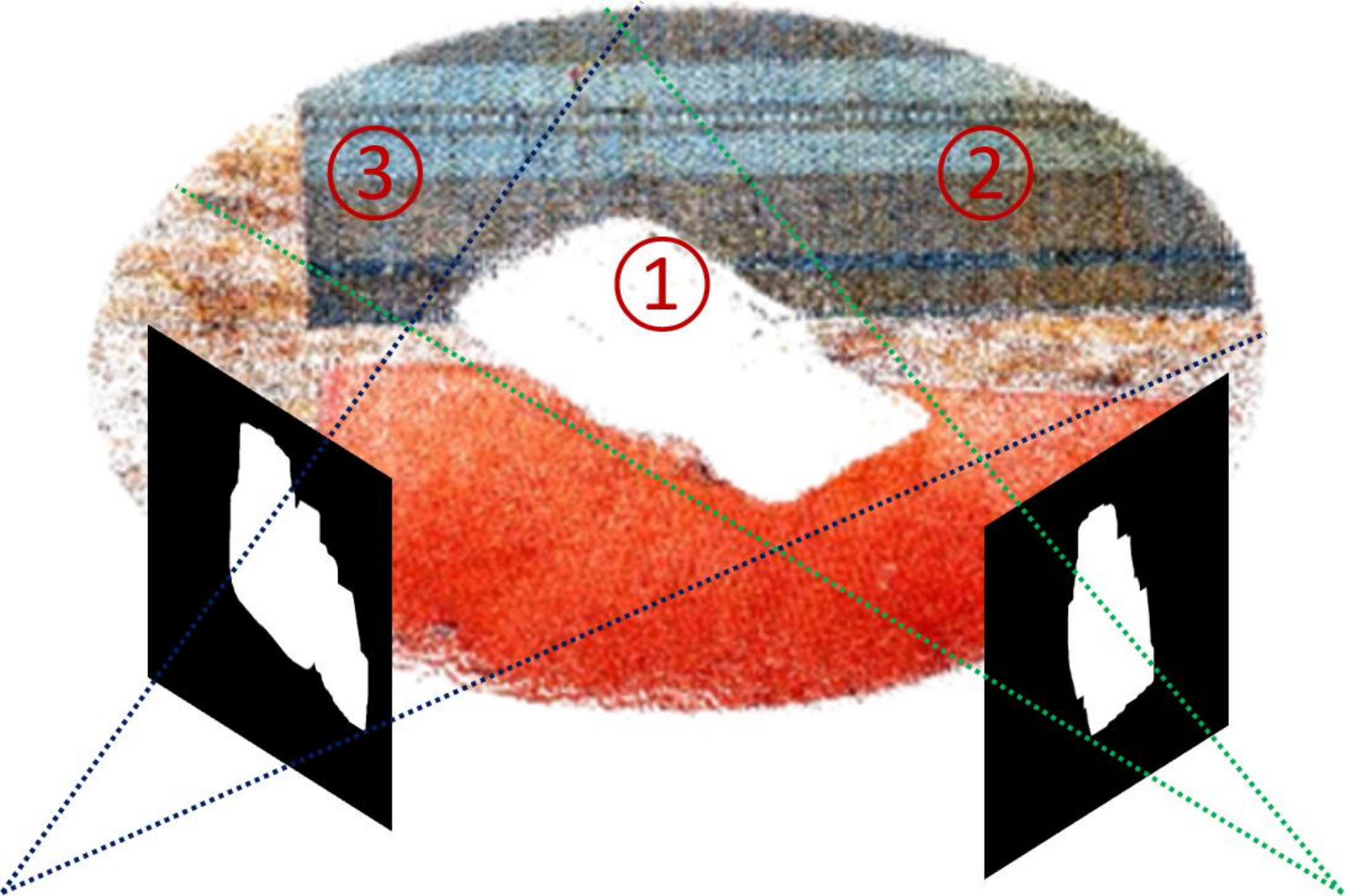} 
        \caption{Mask-remapping interpretation}
    \end{subfigure}
    \begin{subfigure}{0.23\textwidth}
        \centering
        \includegraphics[width=\linewidth]{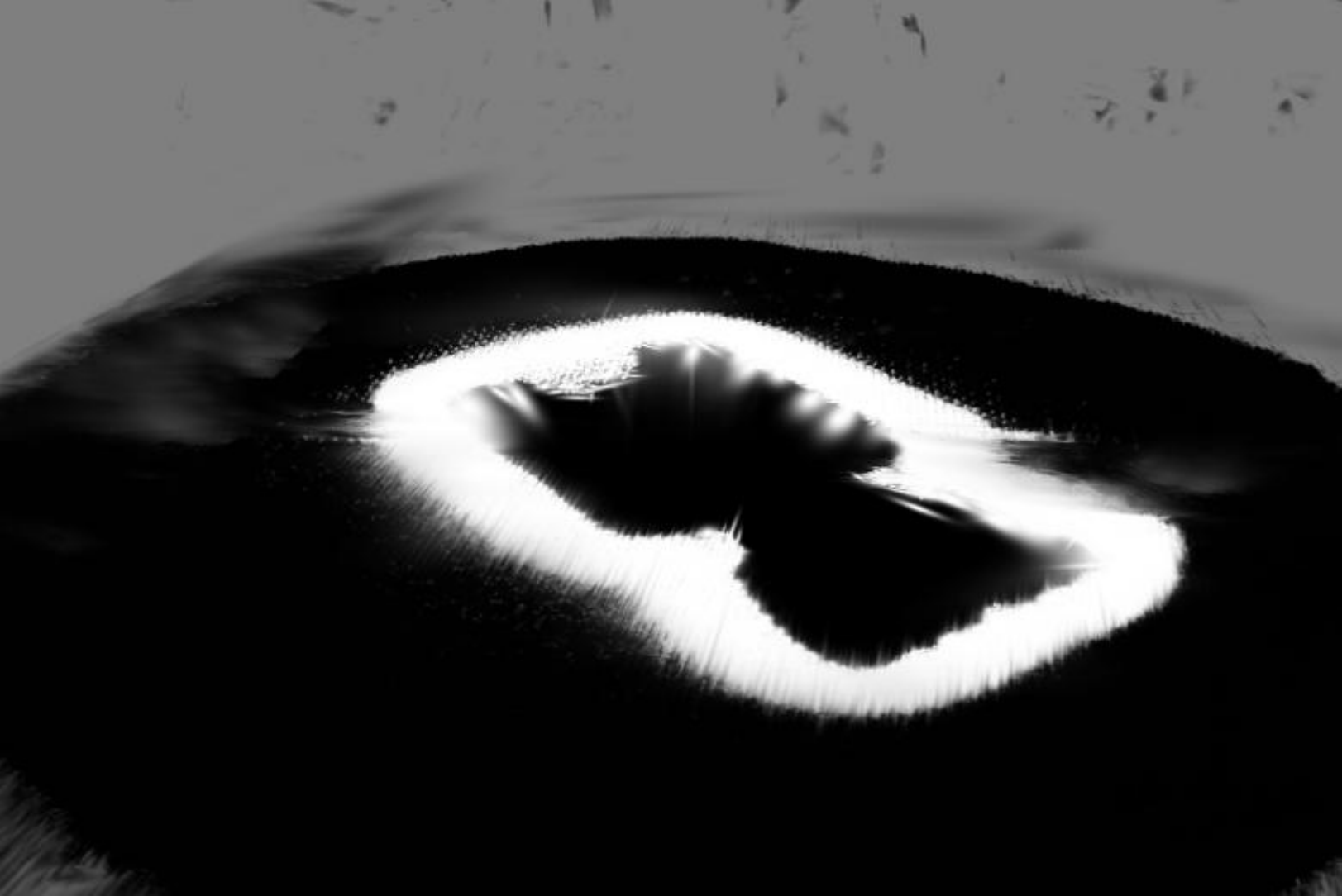} 
        \caption{Renderd mask}
    \end{subfigure}
    \figvspace
    \caption{\textbf{Inpainting mask detection.} (a) shows the object mask (orange) along with the dilated mask (pink), which includes additional neighboring pixels; Each dilated mask consists of two parts: the actual inpainting mask and out-of-interest areas, marked as regions \textcircled{1} and \textcircled{2} in (b). These regions are mapped in 3D to the NBS and background areas, respectively, as shown in regions \textcircled{1} and \textcircled{2} in (c). During multi-view mapping optimization, the central NBS region is progressively refined, while background areas are suppressed by each other; (d) shows the rendered neighbor mask after GS scene optimization.}
    \vspace{-3mm}
    \label{fig:mask and detection}
\end{figure}

\subsection{Inpainting Mask Detection}
\label{sec:additional}

In the overall pipeline of 3D object removal and scene inpainting, identifying the occluded or NBS regions for inpainting is also critical, especially for 360\degree cases. In those cases, the object mask often covers a much larger area than the regions that require inpainting (as illustrated in Fig.~\ref{fig:mask and detection}). Therefore, a novel inpainting mask detection approach is proposed to identify the NBS region for each view.

An intuitive idea is to detect the NBS region in the pruned GS scene and project it back into each view to obtain inpainting masks.
However, since, by definition, the NBS region is occluded by the object, it usually lacks good geometry and meaningful Gaussians, making it challenging to detect in 3D. Here we propose to detect the inpainting masks by leveraging \szh{neighbors of NBS region}. 
In detail, we first dilate the object masks to include neighboring pixels (Fig.~\ref{fig:mask and detection}(a)), and then map the dilated masks onto the pruned scene $\mathcal{G}_{\text{pruned}}$.
Mapping is implemented as updating the learnable parameter $\{p_m\}_{m=1}^M$ of each Gaussian with respect to $\mathcal{L}_1$ loss between rendered masks of $\mathcal{G}_{\text{pruned}}$ and the dilated masks. The optimization process is the same as object removal (Sec.~\ref{sec:remove}) but with the object mask replaced with dilated masks, and the optimization is performed over the pruned scene. We illustrate the \szh{interpretation} of this process in Fig.~\ref{fig:mask and detection}(b-c).
After the mapping process, we are able to get the neighbor mask by rendering $\{p_m\}_{m=1}^M$ from $\mathcal{G}_{\text{pruned}}$ as shown in Fig.~\ref{fig:mask and detection}(d).
Finally, we extract the bounding box for the above rendered mask, which will be taken as input prompt for segment anything model~\cite{kirillov2023segment} to get the final inpainting mask.

\section{Experiments}
\label{sec:experiments}
\noindent\textbf{Datasets:} Most existing publicly available 3D inpainting datasets \cite{mirzaei2023spin, weder2023removing} are limited to front-facing cases, which cannot be used to evaluate methods in more diverse or realistic situations. 
To address this, we construct a dataset that contains 20 diverse cases, ranging from indoor to outdoor, covering \szh{various camera trajectories}-including front-facing, 90°, 180°, and 360°, and containing various interfacial surfaces such as single plane, multiple planes, curvature surfaces, and irregular surfaces. 
Each scene is recorded using a DJI Pocket 3 at $60$fps with a resolution of $1920 \times 1080$, from which we evenly extract 175 images and generate the corresponding camera poses and object masks.
The first 50 images are used for evaluation purpose with the object removed, while the remaining 125 images are object-present and are used to reconstruct the original GS scene.
We use half resolution in the following experiments.
In addition, we evaluate our method on SPINeRF dataset \cite{mirzaei2023spin} to compare with state-of-the-art.

\begin{figure}[t]
\centering
\includegraphics[width=\linewidth]{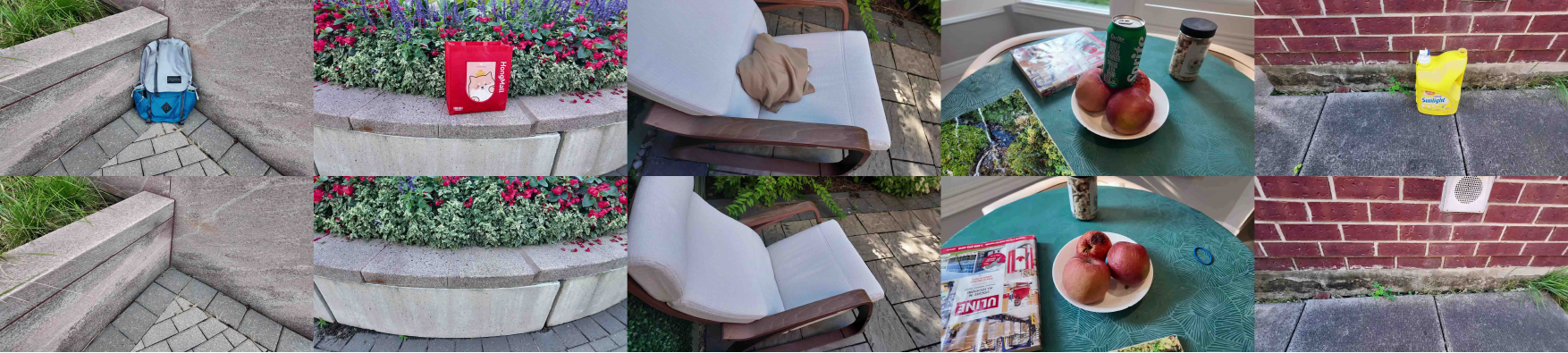}
\includegraphics[width=\linewidth]{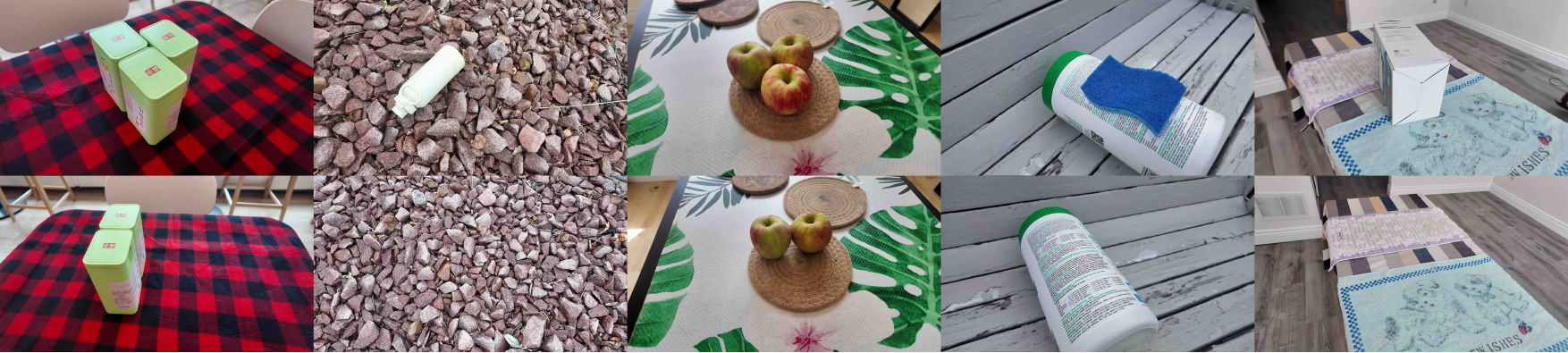} 
\figvspace
\figvspace
\caption{\textbf{Preview of our collected dataset}. The training images are displayed at the top, and the testing images are at the bottom. More visual data about the dataset can be found in the supplement.}
\vspace{-3mm}
\label{fig:dataset}
\end{figure}

\noindent\textbf{Implementation Details:} 
StableDiffusionInpaintingPipeline\footnote{https://huggingface.co/stabilityai/stable-diffusion-2-inpainting} is utilized for image inpainting and our multi-view refinement model initialization. During training, we randomly select $L=5$ images from training images and crop 512$\times$512 patches to fine-tune the model over 1000 iterations with the learning rate and batch size set as $3\times10^{-5}$ and 1, individually.
At inference, directly refining all warped images, i.e., $L=125$ would result in prohibitive memory usage and computational costs. 
To address this, we cluster all views into $L$ groups and select the center view of each group as the anchor. Multi-view refinement is applied only to these $L$ anchor views, while the remaining views in each group are warped from their respective anchors, as artifacts caused by warping between closely aligned views are negligible. 
$L$ is set to 12 for 360\degree cases and linearly scaled down for other scenarios, such as 4 for 90\degree cases.
\szh{We directly use object masks when evaluating on front-facing scenes and apply inpainting mask detection on others.}
All experiments are conducted on a single NVIDIA V100 GPU, with the inpainting of a 3D scene taking approximately one hour.
\begin{table}[t]
	\footnotesize
	\centering
	\caption{\textbf{Quantitative evaluation on the SPINeRF dataset and our dataset.} Numbers in bold indicate the best performance, and underscored numbers indicate the second best.}
    \figvspace
	\begin{adjustbox}{width=\linewidth}
		\begin{tabular}{ccccccc}
			\toprule
			\multicolumn{1}{c}{\multirow{2}*{Method}} 
			&\multicolumn{3}{c}{Ours} 
			&\multicolumn{3}{c}{SPINeRF}\\ 
			\cmidrule(r){2-4} 
			\cmidrule(r){5-7}
			& PSNR ($\uparrow$) & LPIPS ($\downarrow$) & FID ($\downarrow$) & PSNR ($\uparrow$) & LPIPS ($\downarrow$) & FID ($\downarrow$)\\
			\midrule
			GaussianEditor &15.71& 0.6163&  375.03& 14.41& 0.6247& 343.16\\
            GScream &17.18& 0.4431&  290.63& 16.96& \textbf{0.3931}& \underline{154.71}\\
            SPInNeRF & \underline{18.75}&  \underline{0.3519}& \underline{206.43}& 17.47& 0.5740 &239.99\\
            MVIPNeRF & 18.63&  0.4332& 278.99& \textbf{17.67}& 0.5070 &255.51 \\
            \rowcolor{mygray} Ours &\textbf{19.67}& \textbf{0.2685}& \textbf{149.52}& \underline{17.58}& \underline{0.4513} & \textbf{154.34}\\
			\bottomrule
	\end{tabular}
	\end{adjustbox}
	\vspace{-3mm}
	\label{tab:comparison with SOTA}
\end{table}


\begin{figure*}[htbp]
	\scriptsize
	\centering
	\begin{tabular}{cc}
		\hspace{-0.4cm}
        \begin{adjustbox}{valign=t}
			\begin{tabular}{c}
				\includegraphics[width=0.28\textwidth]{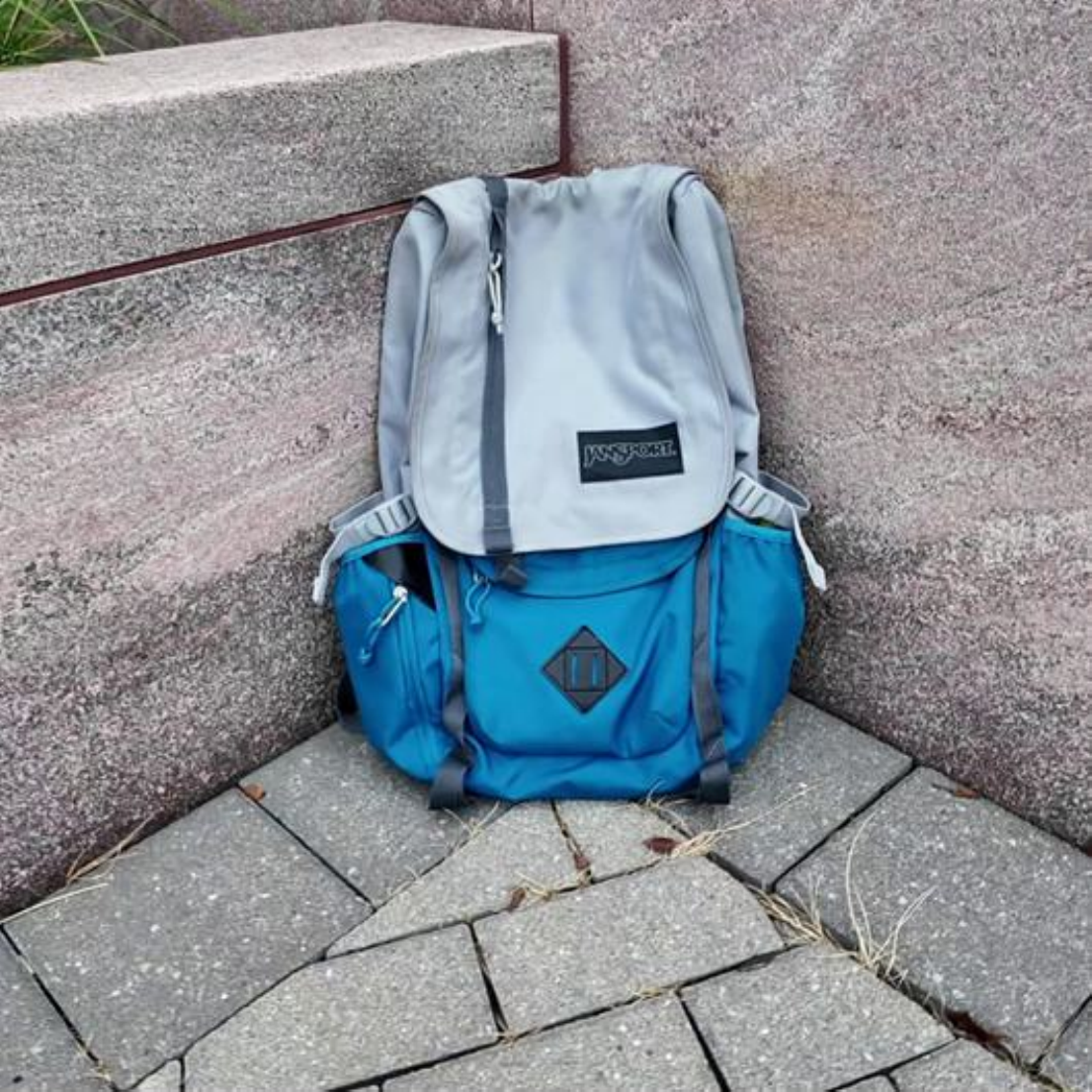}
				\\
			\end{tabular}
		\end{adjustbox}
		\hspace{-4.3mm}
		\begin{adjustbox}{valign=t}
			\begin{tabular}{ccccc}
				\includegraphics[width=\widthscale \textwidth]{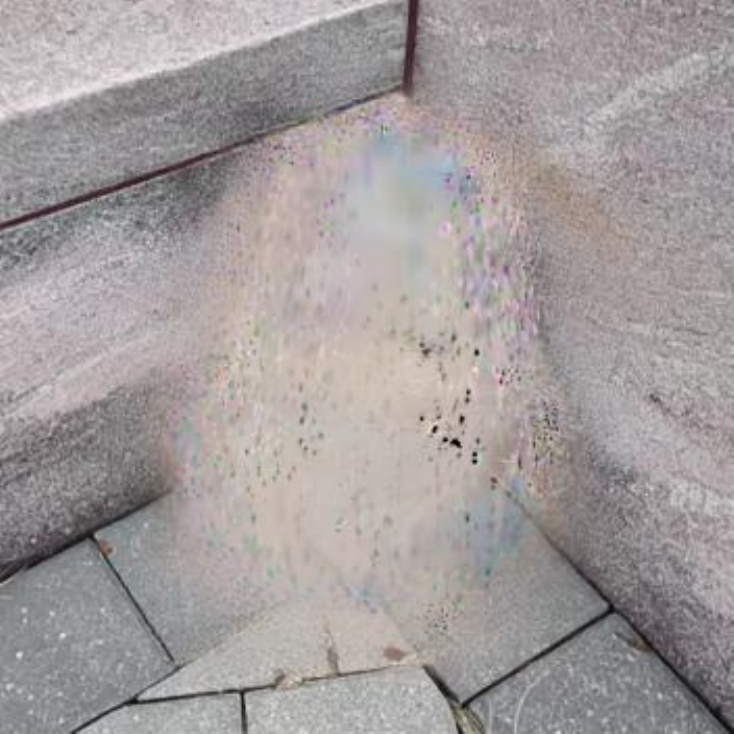} \hspace{-4mm} &
				\includegraphics[width=\widthscale \textwidth]{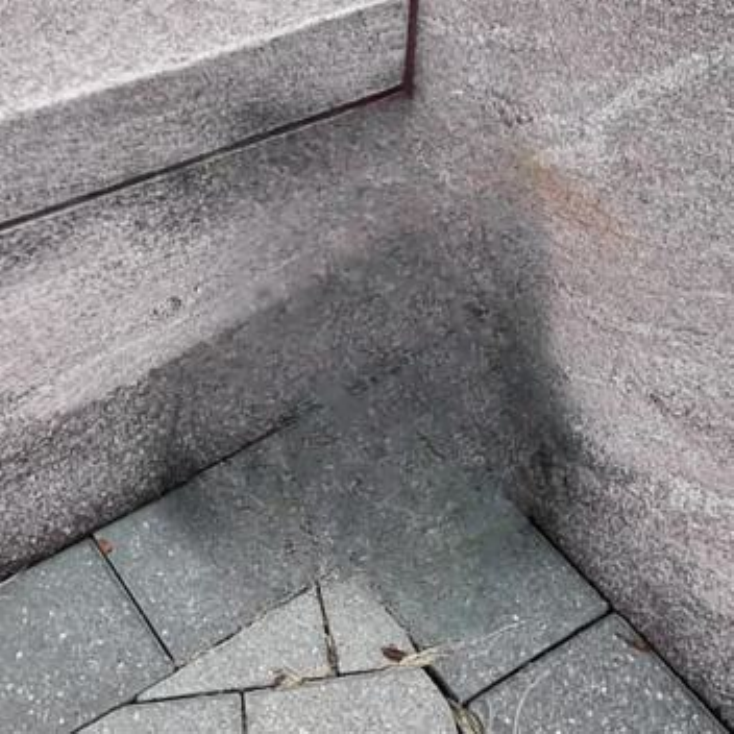} \hspace{-4mm} &
				\includegraphics[width=\widthscale \textwidth]{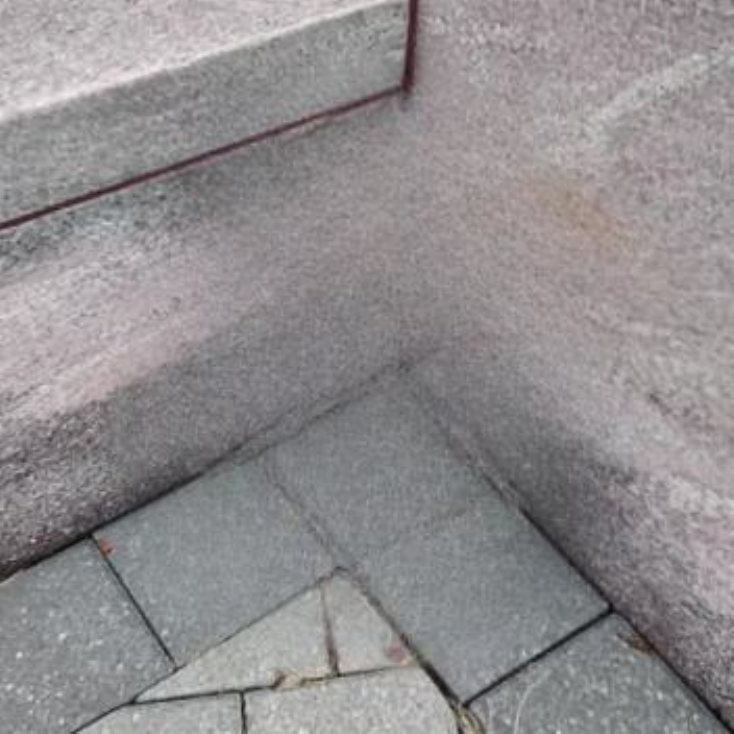} \hspace{-4mm} &
				\includegraphics[width=\widthscale \textwidth]{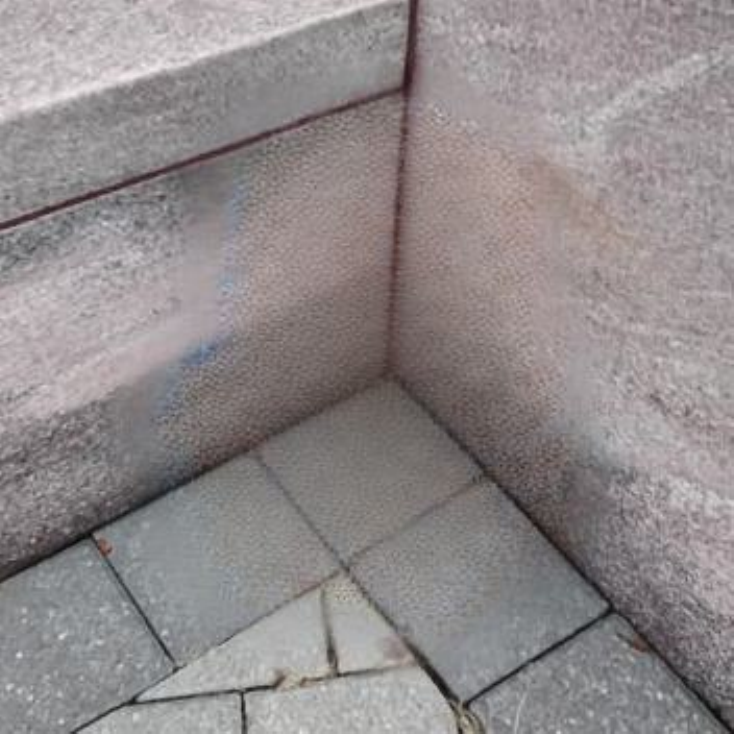}\hspace{-3.5mm} &
				\includegraphics[width=\widthscale \textwidth]{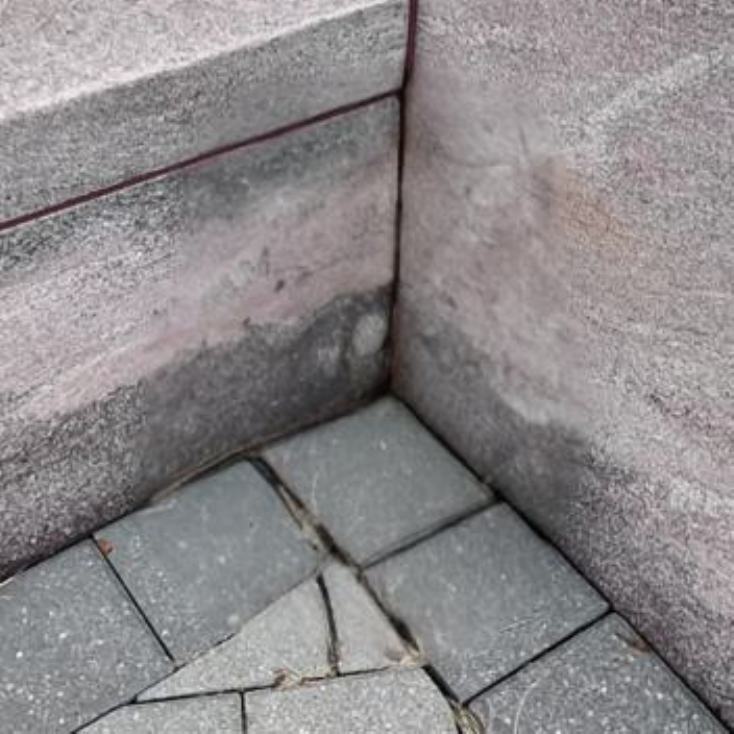}
				\\
                \vspace{-3mm}
				\\
				\includegraphics[width=\widthscale \textwidth]{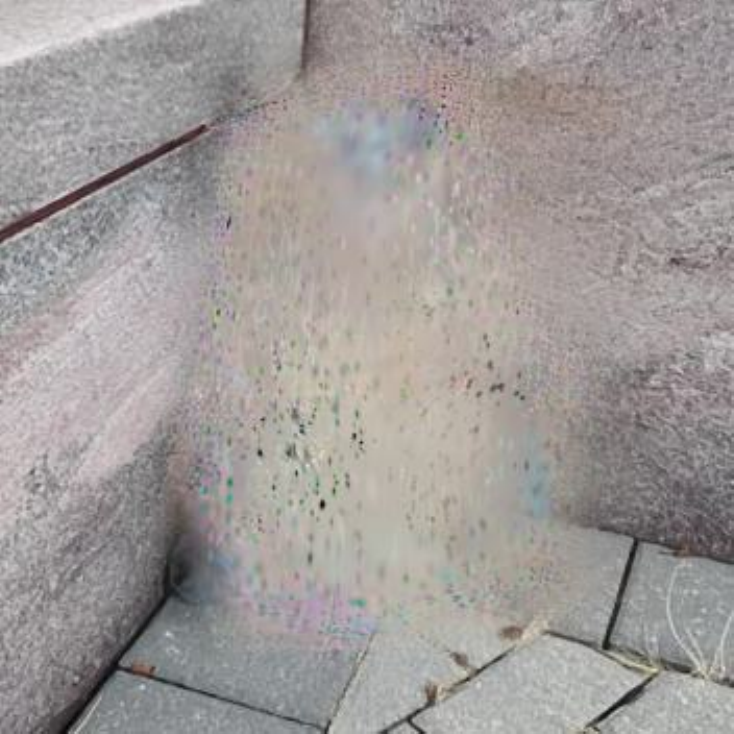} \hspace{-4mm} &
				\includegraphics[width=\widthscale \textwidth]{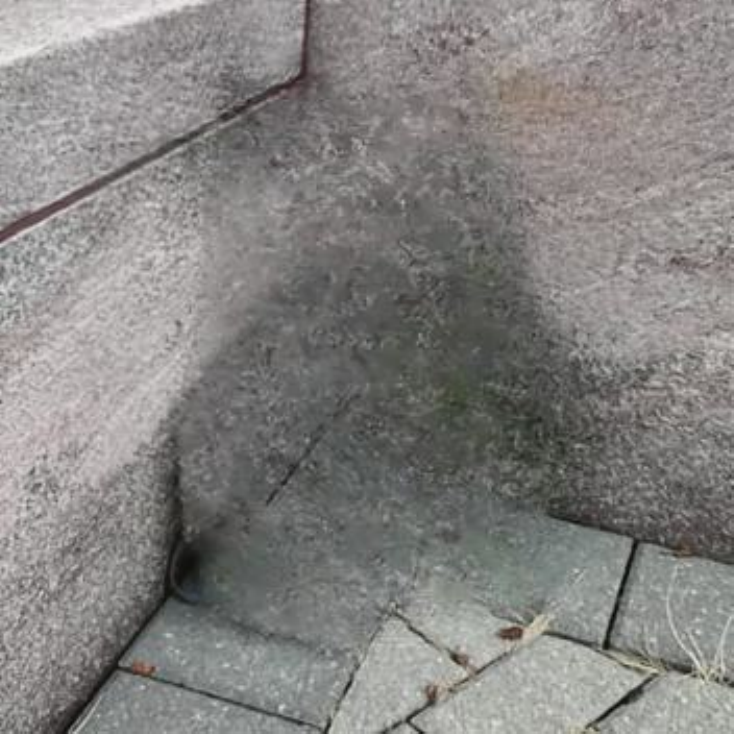} \hspace{-4mm} &
				\includegraphics[width=\widthscale \textwidth]{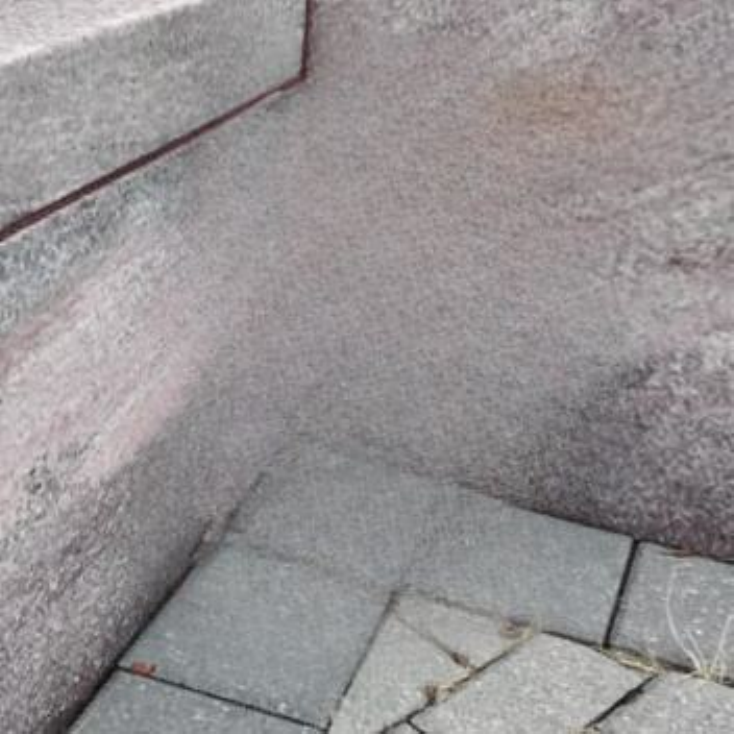} \hspace{-4mm} &
				\includegraphics[width=\widthscale \textwidth]{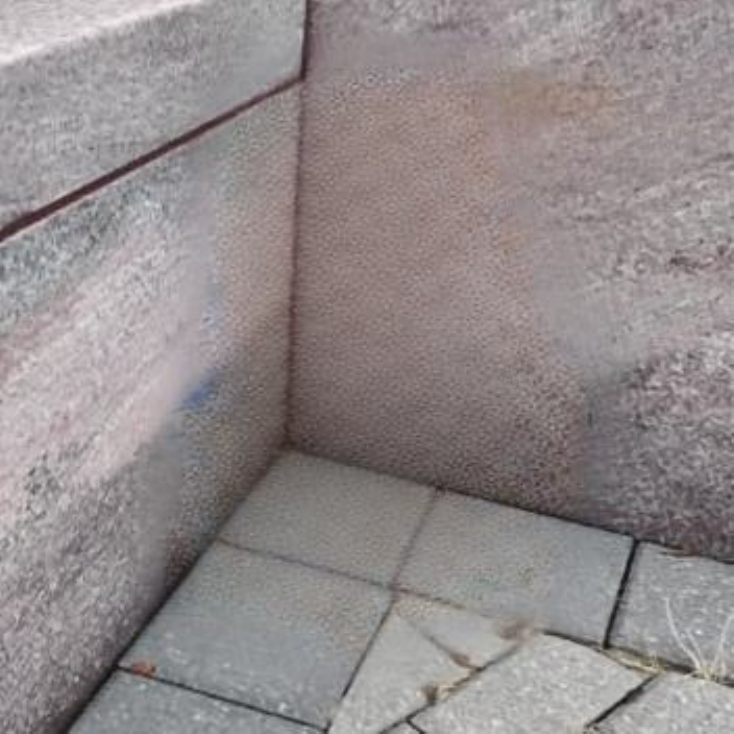}\hspace{-3.5mm} &
				\includegraphics[width=\widthscale \textwidth]{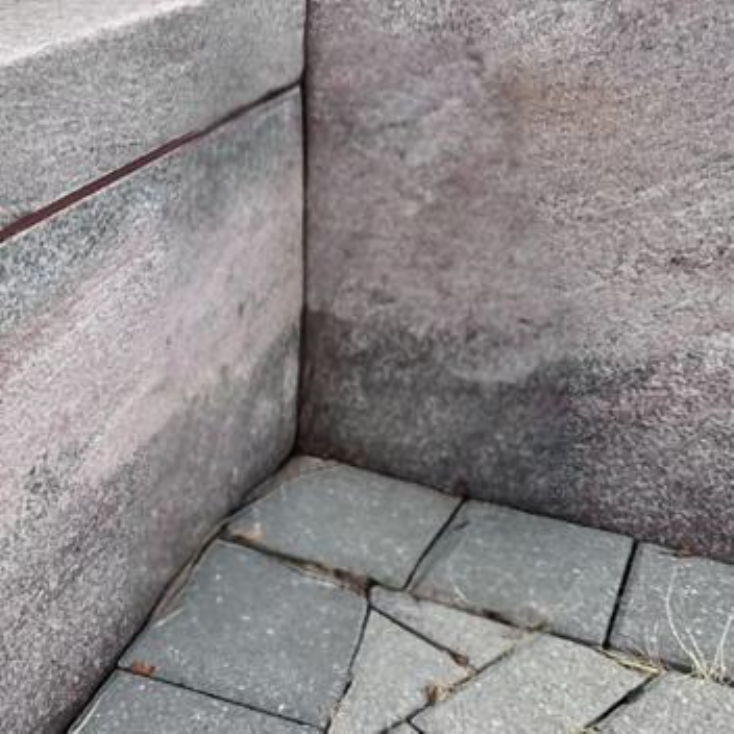}
				\\
    
			\end{tabular}
			\end{adjustbox}

         \\
        \hspace{-0.4cm}
         \begin{adjustbox}{valign=t}
			\begin{tabular}{c}
				\includegraphics[width=0.28\textwidth]{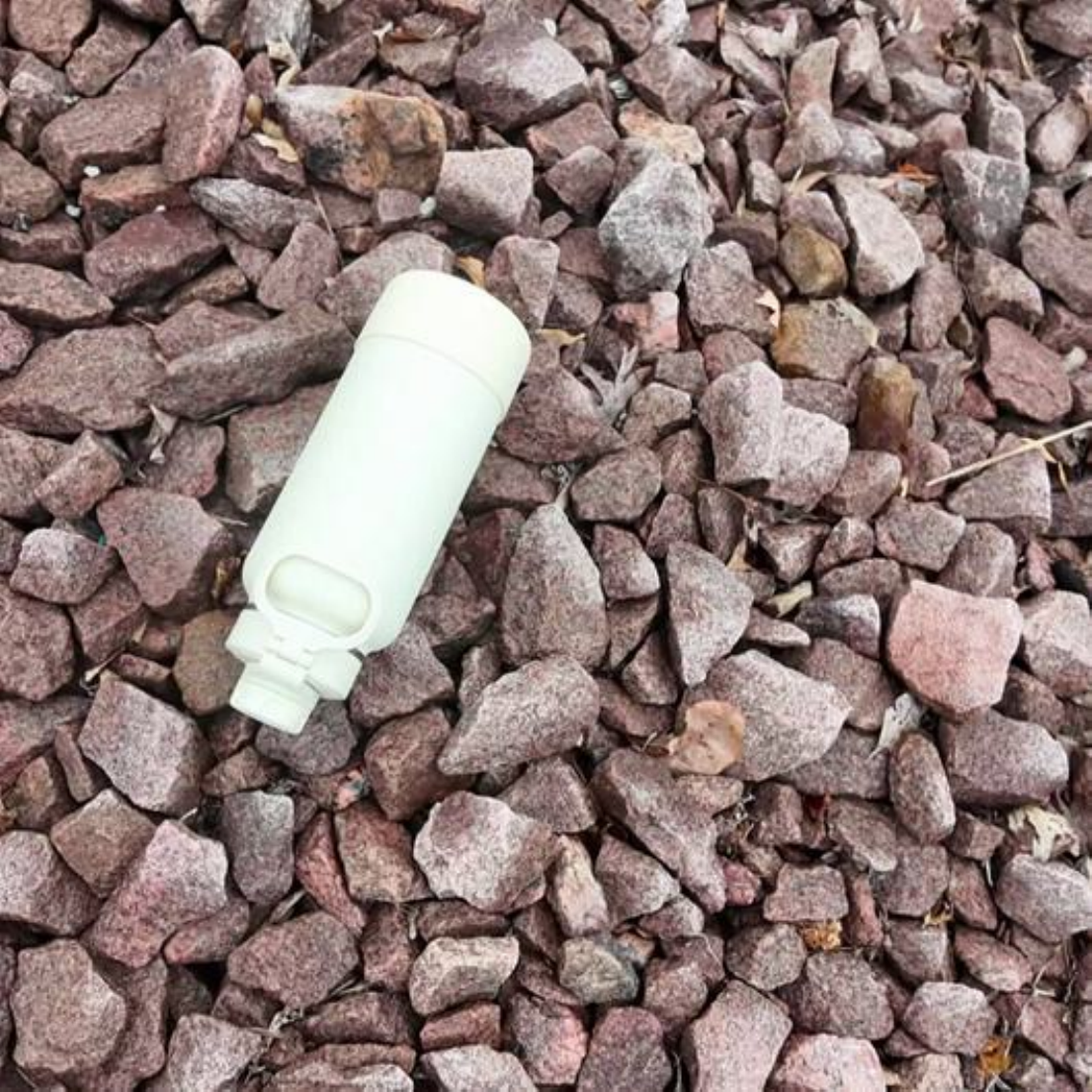}
				\\
			\end{tabular}
		\end{adjustbox}
		\hspace{-4.3mm}
		\begin{adjustbox}{valign=t}
			\begin{tabular}{ccccc}
				\includegraphics[width=\widthscale \textwidth]{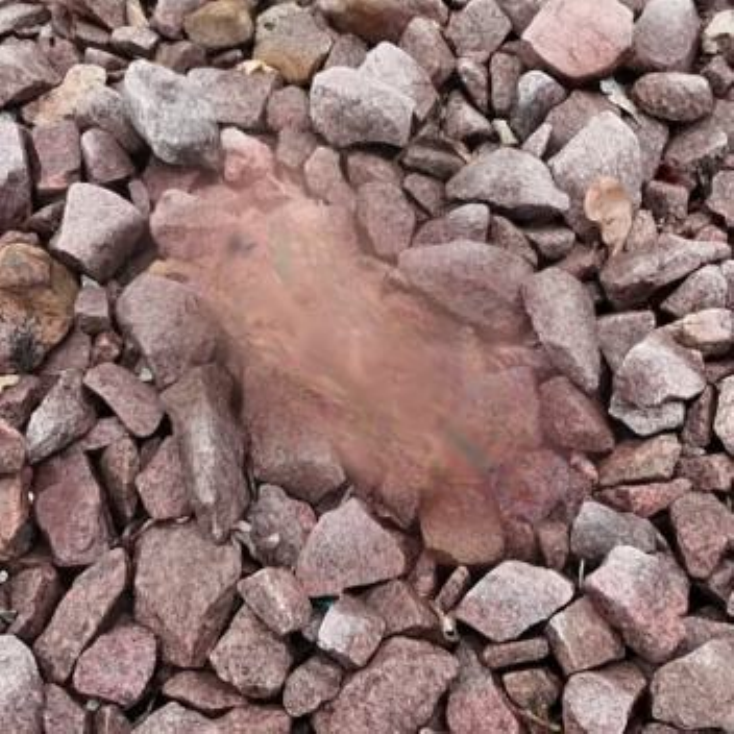} \hspace{-4mm} &
				\includegraphics[width=\widthscale \textwidth]{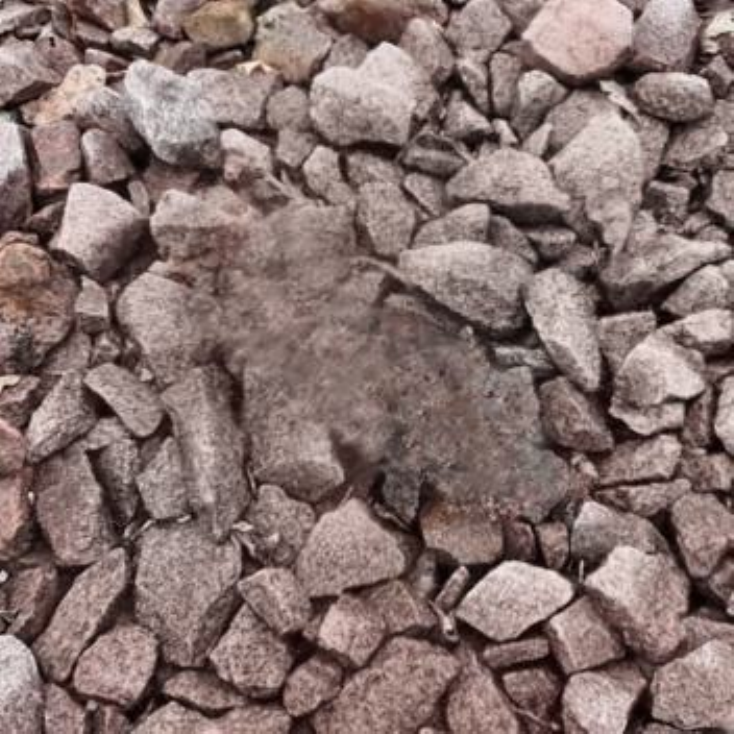} \hspace{-4mm} &
				\includegraphics[width=\widthscale \textwidth]{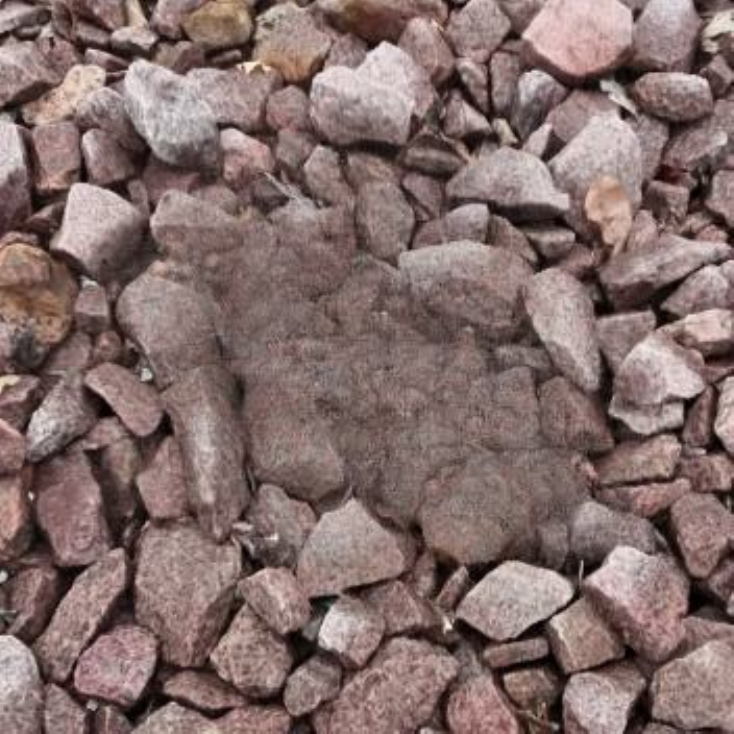} \hspace{-4mm} &
				\includegraphics[width=\widthscale \textwidth]{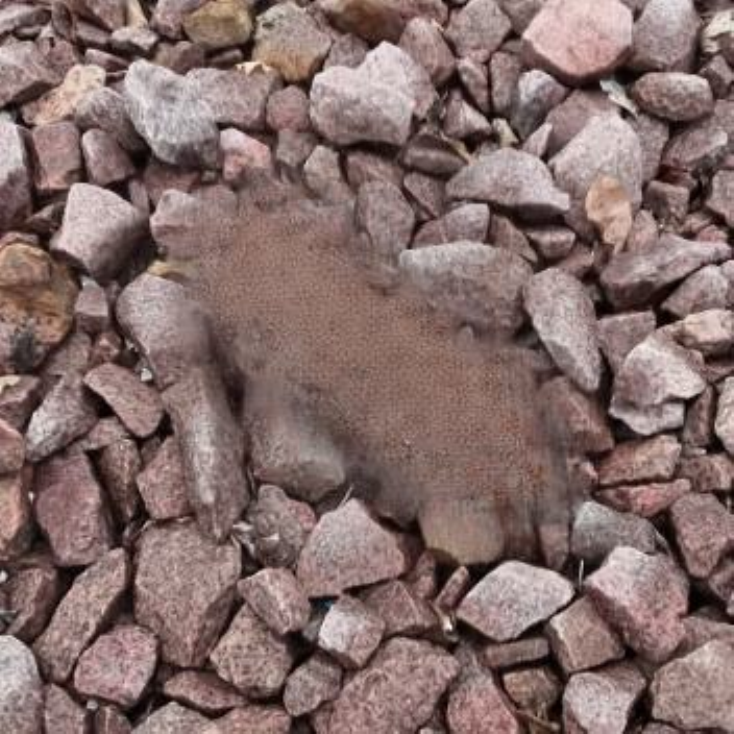}\hspace{-3.5mm} &
				\includegraphics[width=\widthscale \textwidth]{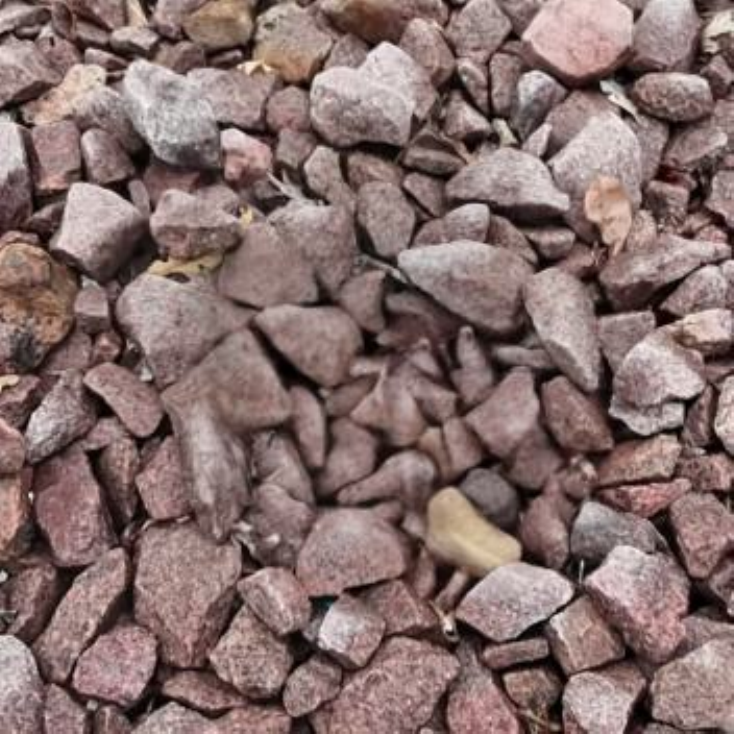}
				\\
                \vspace{-3mm}
				\\
				\includegraphics[width=\widthscale \textwidth]{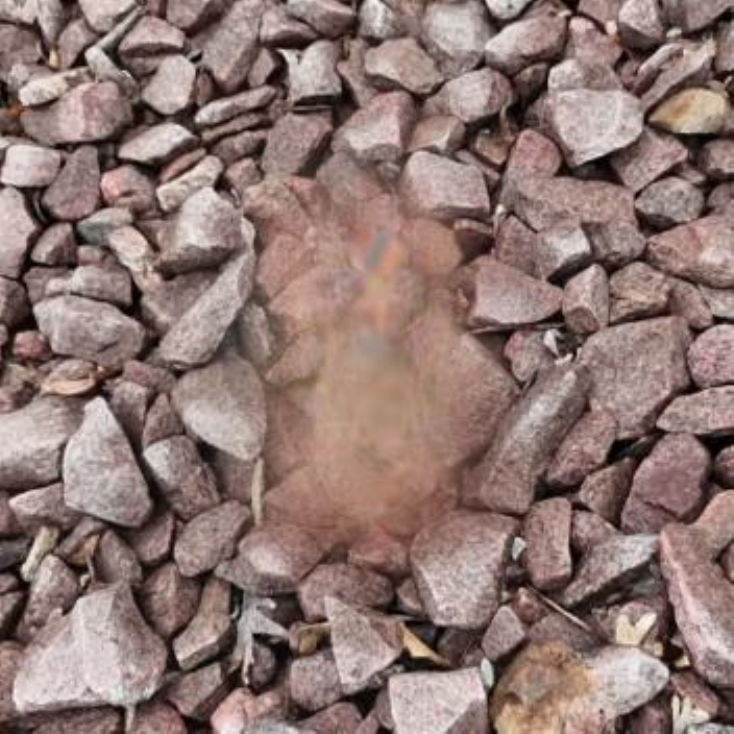} \hspace{-4mm} &
				\includegraphics[width=\widthscale \textwidth]{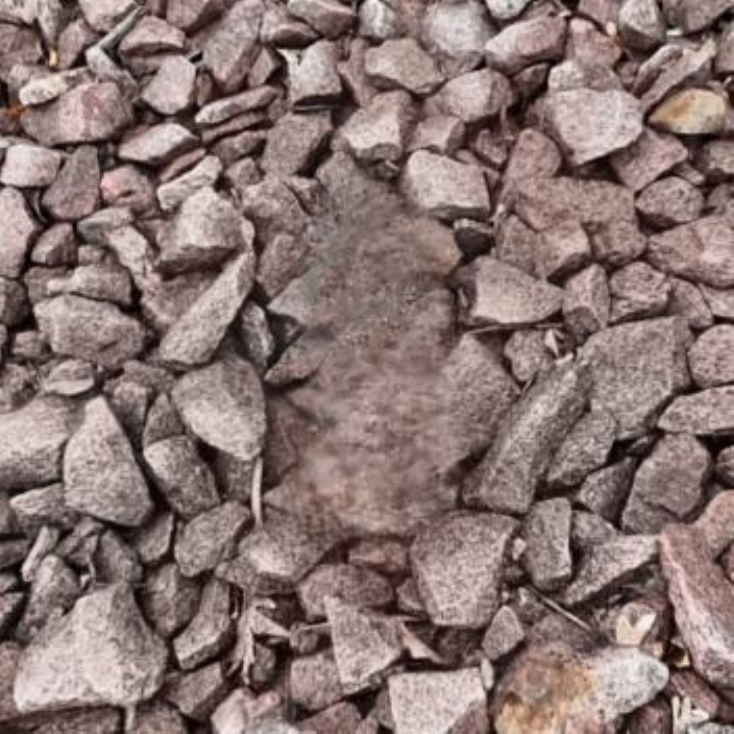} \hspace{-4mm} &
				\includegraphics[width=\widthscale \textwidth]{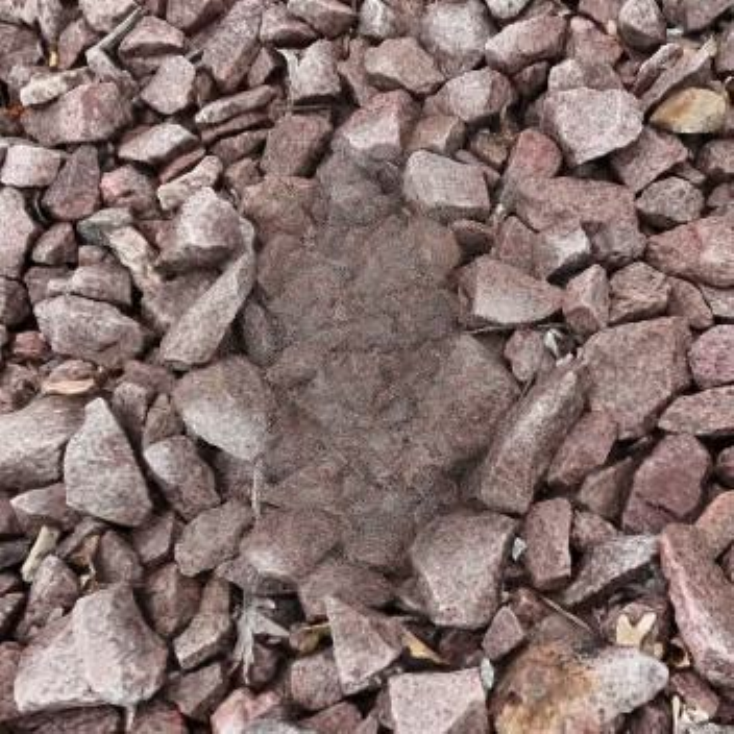} \hspace{-4mm} &
				\includegraphics[width=\widthscale \textwidth]{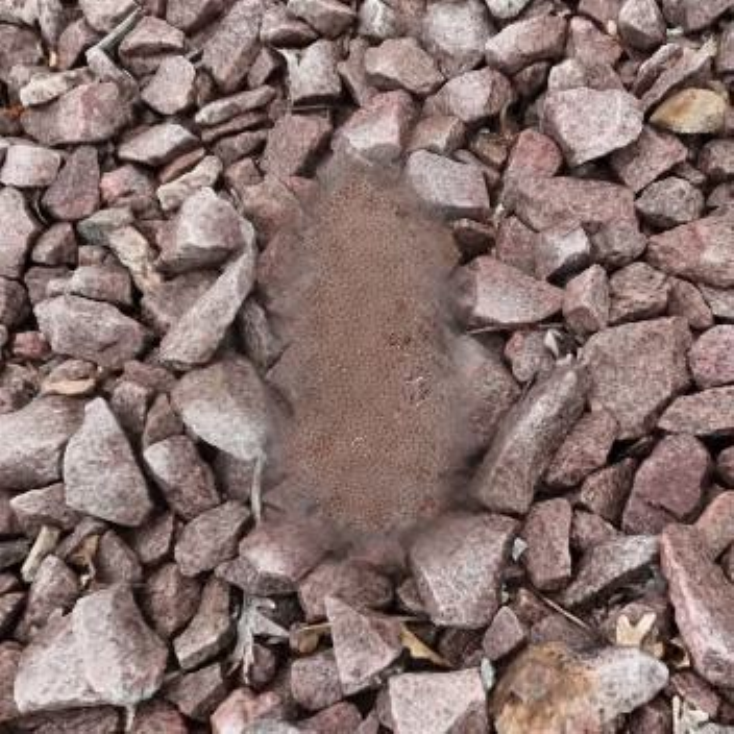}\hspace{-3.5mm} &
				\includegraphics[width=\widthscale \textwidth]{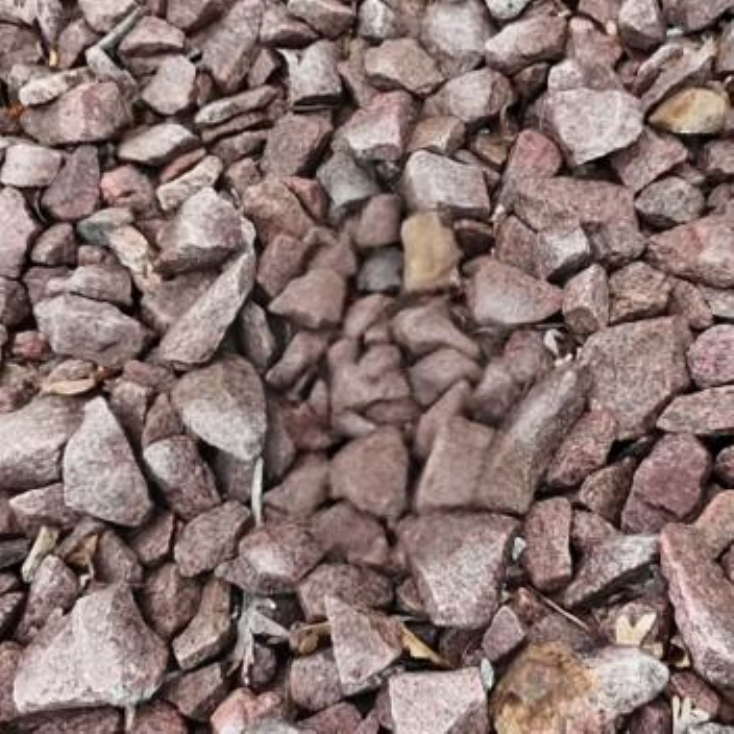}
				\\
    
			\end{tabular}
			\end{adjustbox}
   
					      \\
    
			
  		\hspace{-0.4cm}
         \begin{adjustbox}{valign=t}
			\begin{tabular}{c}
				\includegraphics[width=0.28\textwidth]{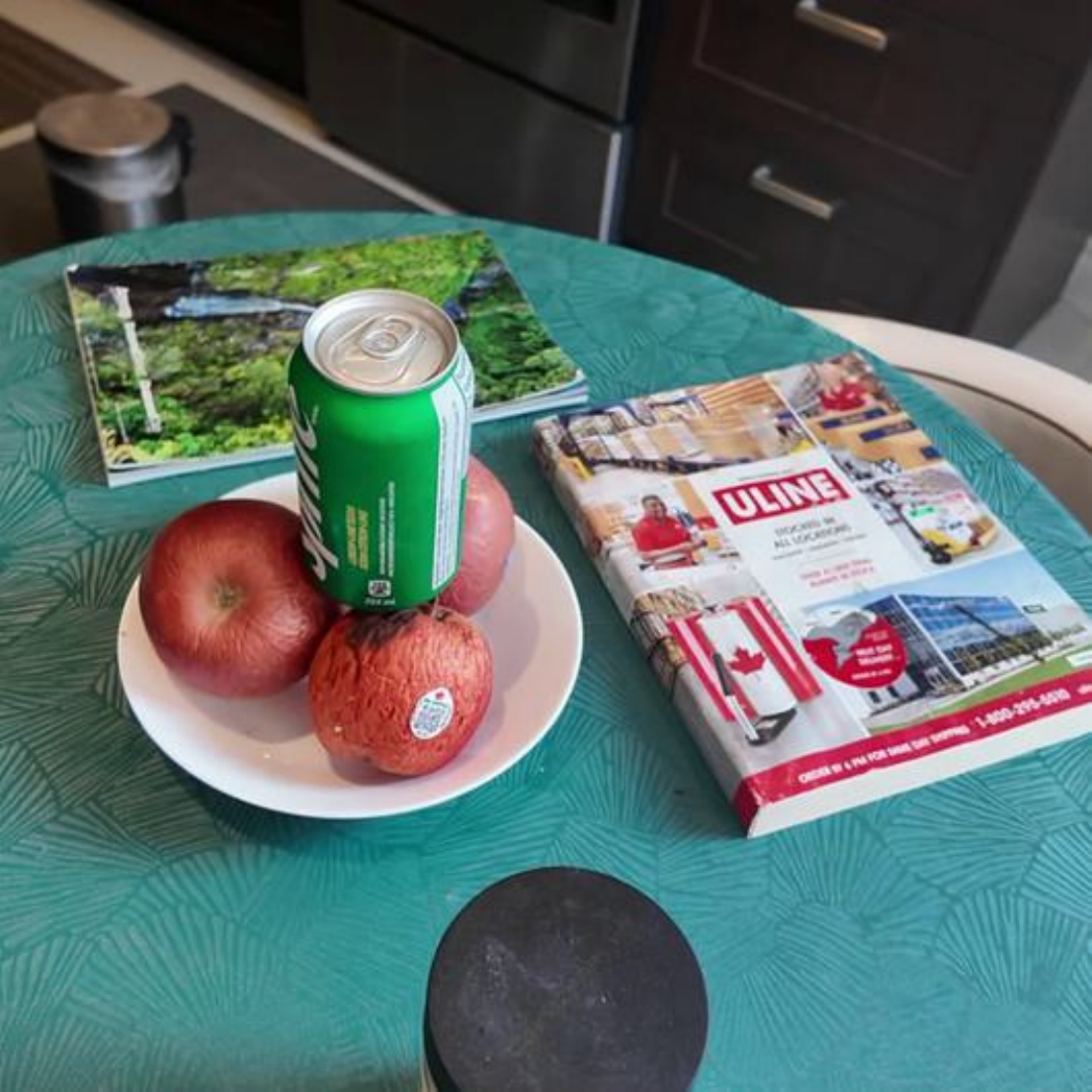}
				\\
			    Input
			\end{tabular}
		\end{adjustbox}
		\hspace{-4.3mm}
		\begin{adjustbox}{valign=t}
			\begin{tabular}{ccccc}
				\includegraphics[width=\widthscale \textwidth]{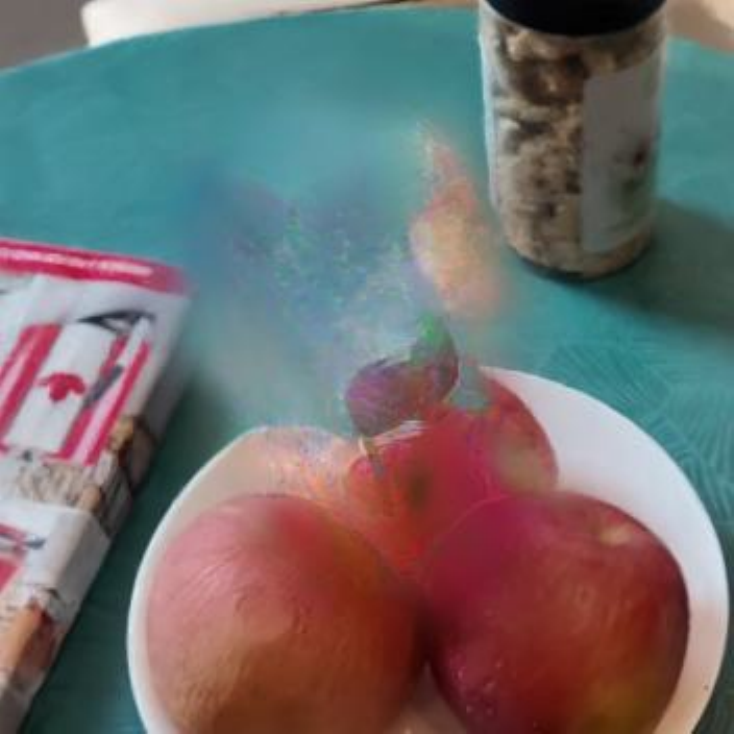} \hspace{-4mm} &
				\includegraphics[width=\widthscale \textwidth]{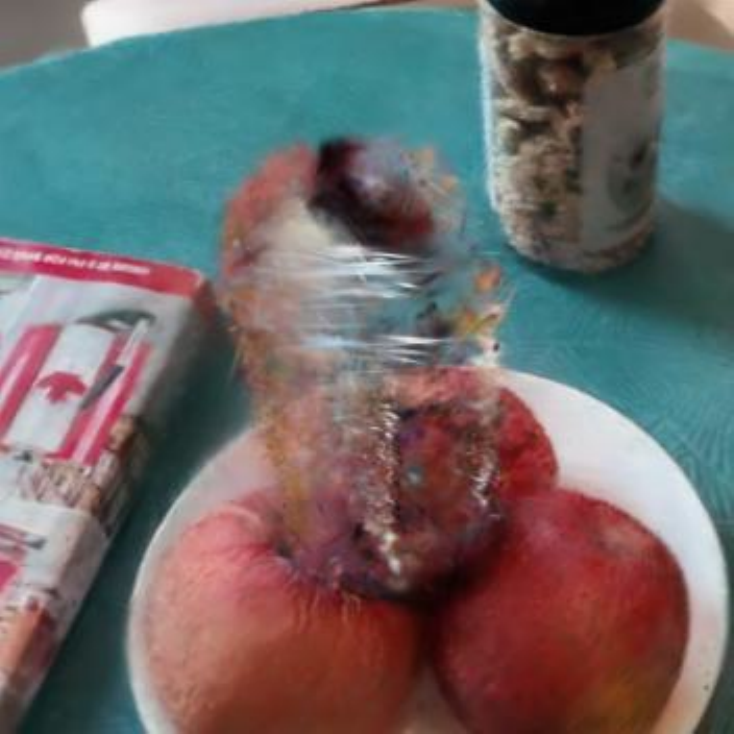} \hspace{-4mm} &
				\includegraphics[width=\widthscale \textwidth]{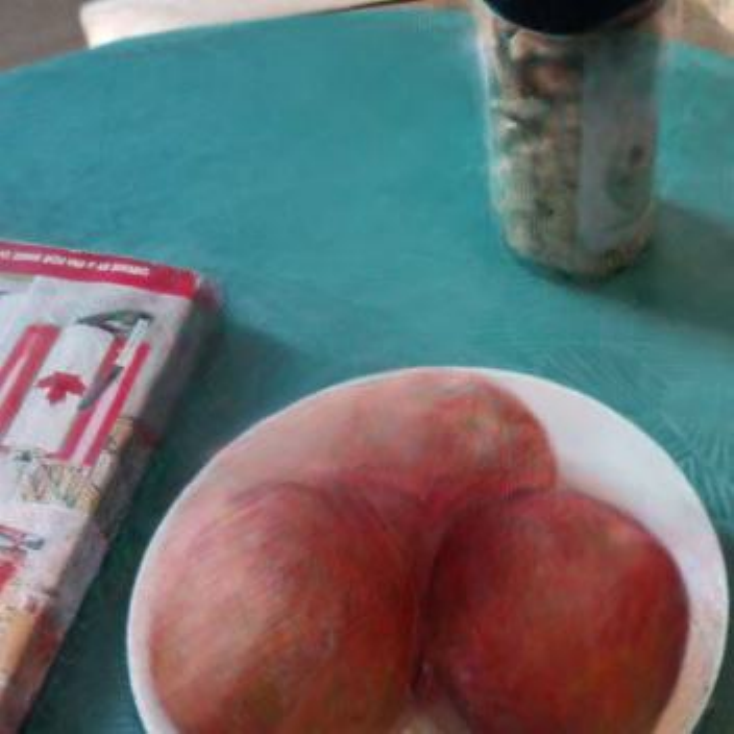} \hspace{-4mm} &
				\includegraphics[width=\widthscale \textwidth]{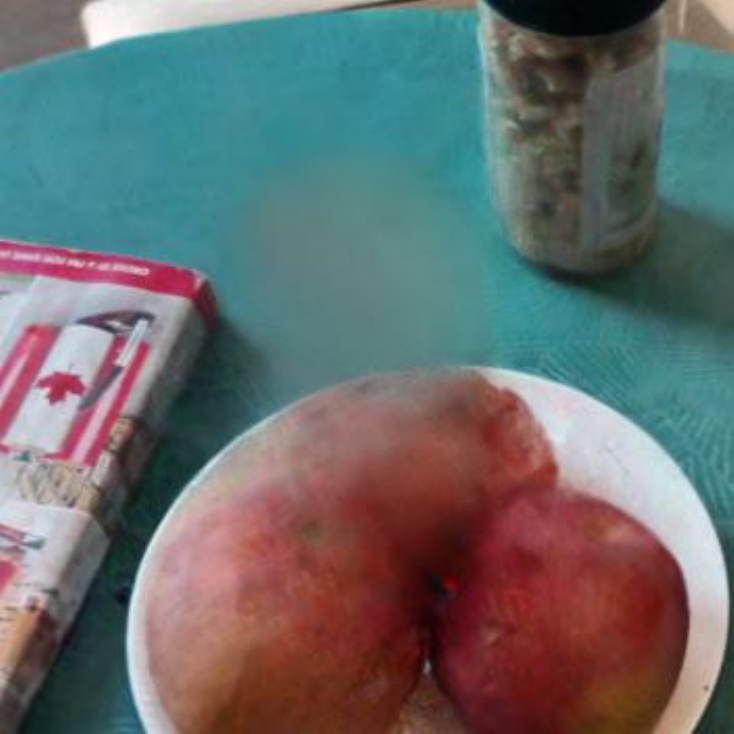}\hspace{-3.5mm} &
				\includegraphics[width=\widthscale \textwidth]{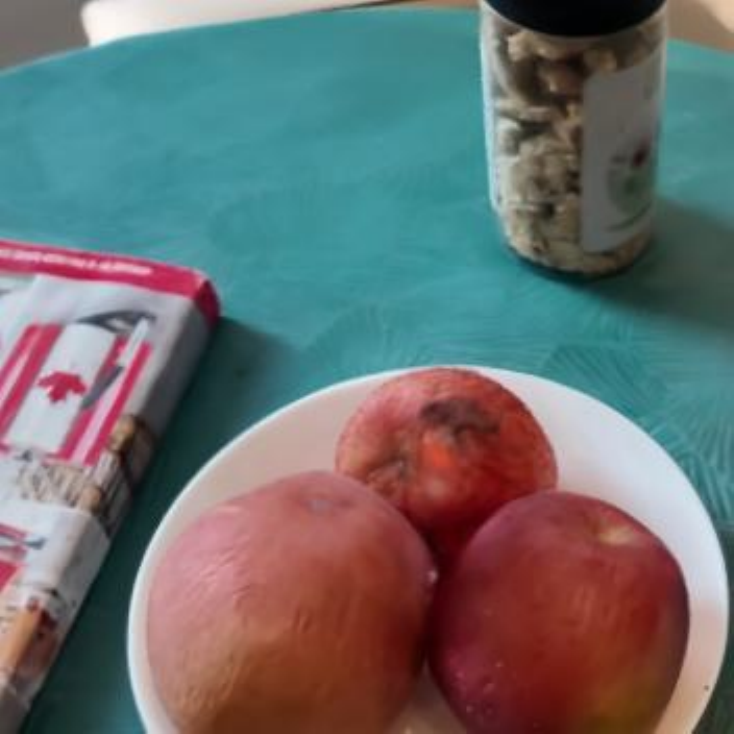}
				\\
                \vspace{-3mm}
				\\
				\includegraphics[width=\widthscale \textwidth]{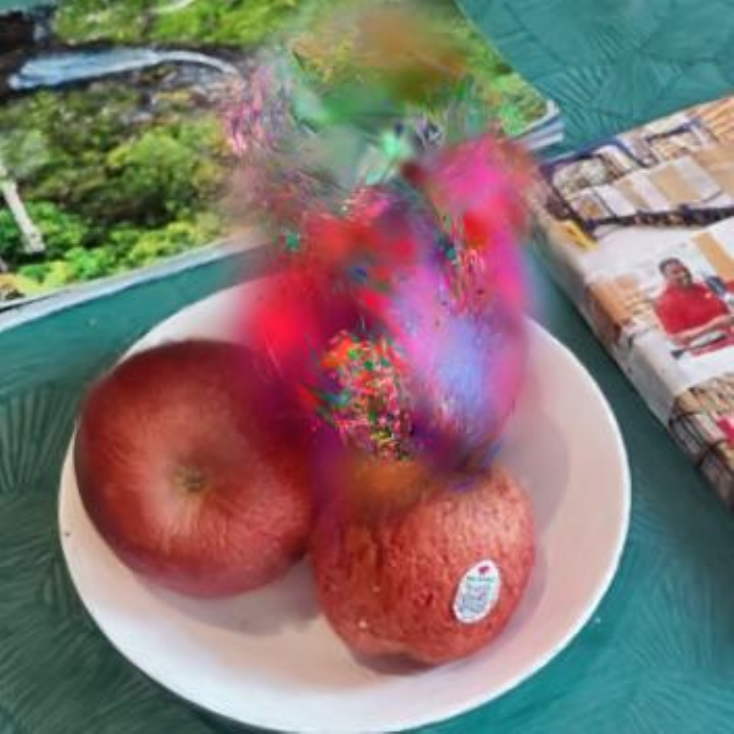} \hspace{-4mm} &
				\includegraphics[width=\widthscale \textwidth]{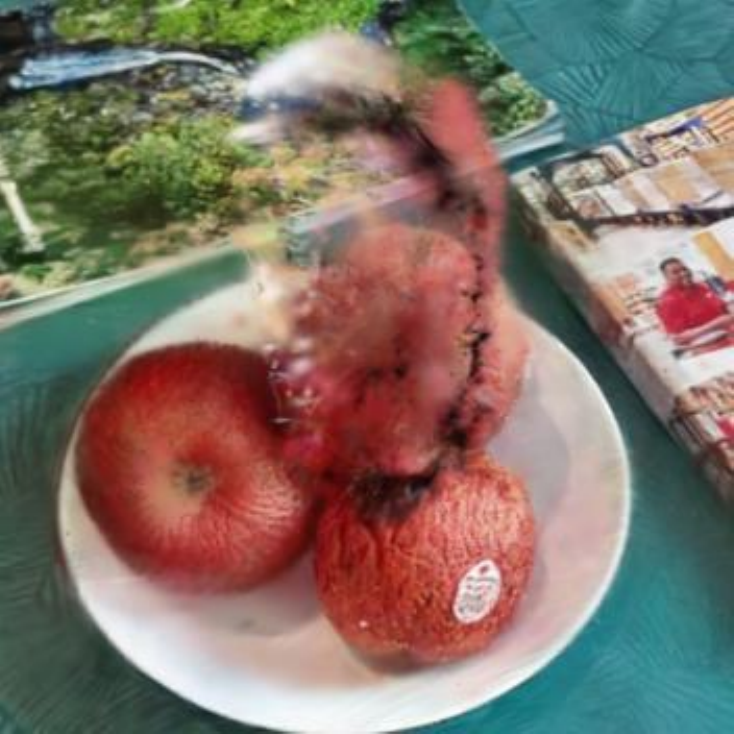} \hspace{-4mm} &
				\includegraphics[width=\widthscale \textwidth]{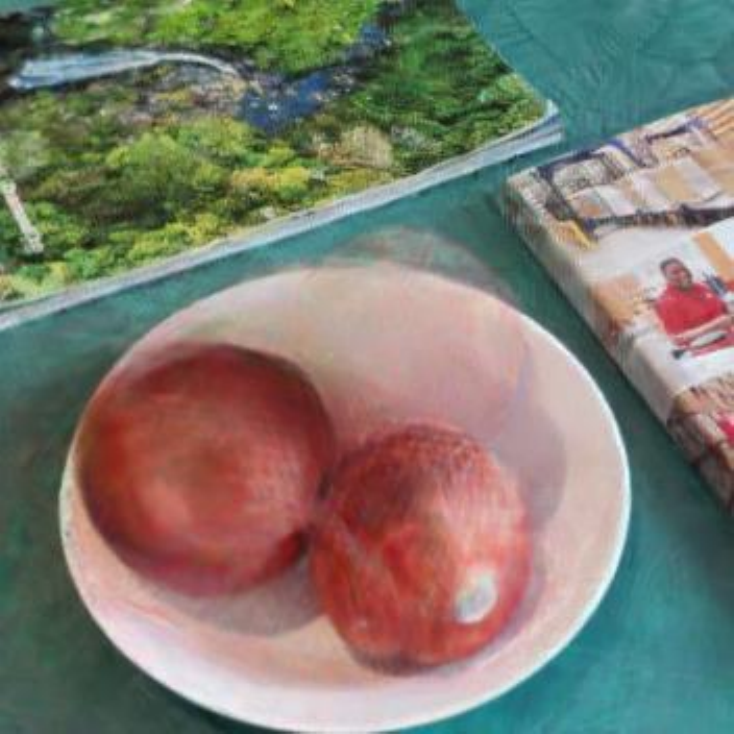} \hspace{-4mm} &
				\includegraphics[width=\widthscale \textwidth]{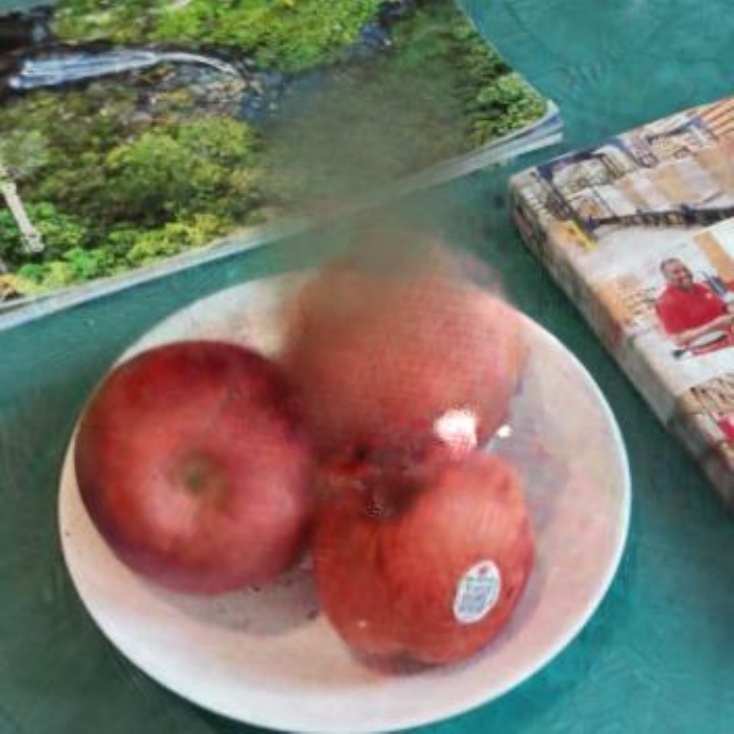}\hspace{-3.5mm} &
				\includegraphics[width=\widthscale \textwidth]{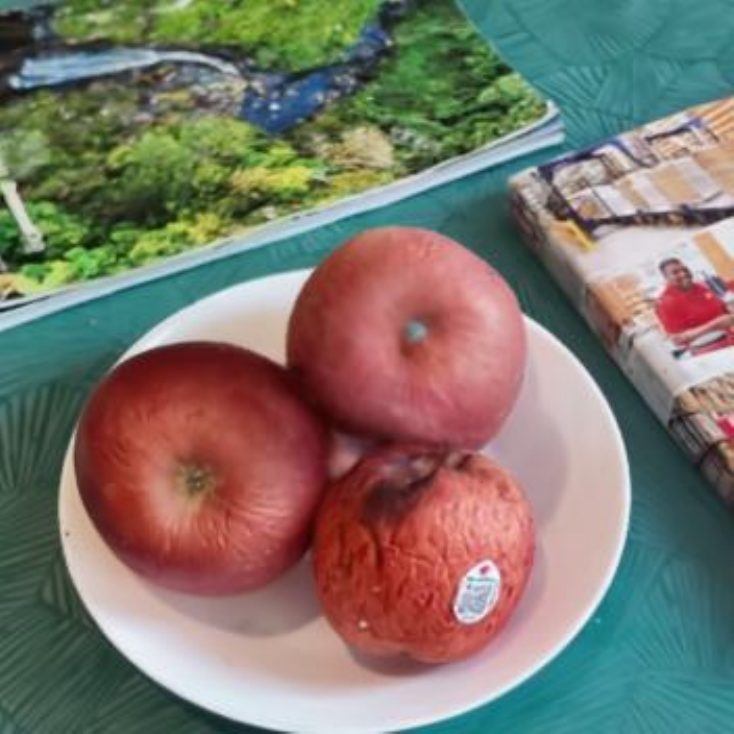}
				\\
    			    GaussianEditor \hspace{-4mm} &
        			GSream \hspace{-4mm} &
    				SPInNeRF \hspace{-4mm} &
    				MVIPNeRF\hspace{-4mm} &
    				\textbf{Ours}
    
				\\
			\end{tabular}
			\end{adjustbox}
  
	\end{tabular}
    \figvspace
	\caption{\textbf{Qualitative evaluation.} We show the visual comparisons against the state-of-the-art 3D inpainting algorithms, with two views displayed at the top and bottom for each method. Our method generates consistent, higher-quality results with rich texture details and fewer artifacts.}
	\label{fig:qualitative comparisons}
	\vspace{-5mm}
\end{figure*}

\subsection{Evaluation against State-of-the-Art Methods}
We evaluate our method against the state-of-the-art methods on our dataset and SPINeRF dataset.
The following state-of-the-art methods are reproduced on on our datasets, including GaussianEditor~\cite{chen2024gaussianeditor}, GScream~\cite{wang2024gscream}, MVIPNeRF~\cite{chen2024mvip} and SPInNeRF~\cite{mirzaei2023spin} using their official implementations. 
In particular, since the naive NeRF models used in MVIPNeRF and SPInNeRF cannot handle complex unbounded scenes, we replace them with MeRF~\cite{reiser2023merf} considering both memory usage and reconstruction performance. 
On our dataset, we utilize the evaluation protocol and official implementation from \cite{chen2024mvip} to compare methods. Specifically, we render color images at corresponding test views from the inpainted scenes and evaluate them against the ground truth using three metrics, namely PSNR, LPIPS, and FID. These metrics are calculated only within the bounding box defined by the object mask. For the SPINeRF dataset, we copy values from \cite{chen2024mvip} for SPInNeRF and MVIPNeRF, and we reproduce GaussianEditor and GScream to align with the metric computation protocol in \cite{chen2024mvip}.

As can be seen from Tab.~\ref{tab:comparison with SOTA}, the proposed method consistently achieves the best performance on our dataset, with a particularly large margin that demonstrates its superiority in unconstrained realistic scenarios. 
On the SPINeRF dataset, our method achieves state-of-the-art results comparable to other approaches, given that this dataset is limited to front-facing cases and challenging to distinguish methods in handling complex cases.

In addition, we provide qualitative comparisons in Fig.~\ref{fig:qualitative comparisons}, where our method successfully removes objects and seamlessly inpaints with appropriate content. Moreover, the inpainted scene maintains 3D consistency, yielding visually more pleasing outcomes and a coherent geometric structure. In contrast, other methods often produce severe artifacts and fail to inpaint objects effectively in complex, realistic scenarios. We highly recommend viewing our video visualizations for a more comprehensive comparison.

\subsection{Detection of Inpainting Masks}
We then demonstrate the importance of doing inpainting mask detection and differentiating them from object masks. For our collected dataset, we manually annotate inpainting masks from rendered images of pruned scenes and compare them with the original object masks. Since our dataset is derived from continuous capture videos, we annotate one out of every 10 images in each scene, resulting in a total of 260 annotated masks. Our analysis shows that the inpainting masks, on average, occupy only 50.78\% of the area covered by the object masks. This finding highlights that adopting entire object masks requires inpainting nearly twice the necessary area, complicating the task and leading to suboptimal results.

We also compare our detected inpainting masks with the manually labeled masks using the mean Intersection over Union (mIoU) metric. Our method achieves an 81.12\% IoU score, validating the accuracy and robustness of the proposed inpainting mask detection strategy. In contrast, the method proposed in \cite{chen2024gaussianeditor} achieves only 42.55\%, which is significantly inferior to our approach.

\subsection{Ablation Study}
Our dataset is chosen for the ablation study due to its diversity and complexity.

\begin{table}[t]
    \centering
    \footnotesize
    \caption{\textbf{Quantatitive results of ablation studies.}}
    \figvspace
    \begin{tabular}{cccc}
        \toprule
        Method & PSNR ($\uparrow$) & LPIPS ($\downarrow$) & FID ($\downarrow$)\\
        \midrule
        w/o Warping & 17.85& 0.3215& 198.24 \\
        \bottomrule
        w/o Refinement & 18.90 &0.3069 &206.96 \\
        General Refinement & 19.08 &0.2719 &165.80 \\
        Single-View Refinement & 19.46& 0.2725& 154.33\\
        \rowcolor{mygray}Multi-View Refinement (Ours) &19.67& 0.2685&  149.52 \\
        \bottomrule
    \end{tabular}
    \vspace{-4mm}
    \label{tab:ablation}
\end{table}

\noindent \textbf{Effectiveness of geometry-guided warping:}
With image rendering of the pruned scene and inpainting masks, we adopt a warping-refinement scheme rather than directly inpainting the masked areas. To demonstrate the effectiveness of our approach, we disable geometry reconstruction and warping operations, repurposing the multi-view refinement model as a multi-view inpainting model. Specifically, we zero out the masked regions in the input images and attempt to consistently inpaint the missing areas. The inpainting model is also fine-tuned for each given scene, with all other settings remaining unchanged.

\begin{figure}[h]
    \centering    
    \begin{subfigure}{0.15\textwidth}
        \centering
        \includegraphics[width=\linewidth]{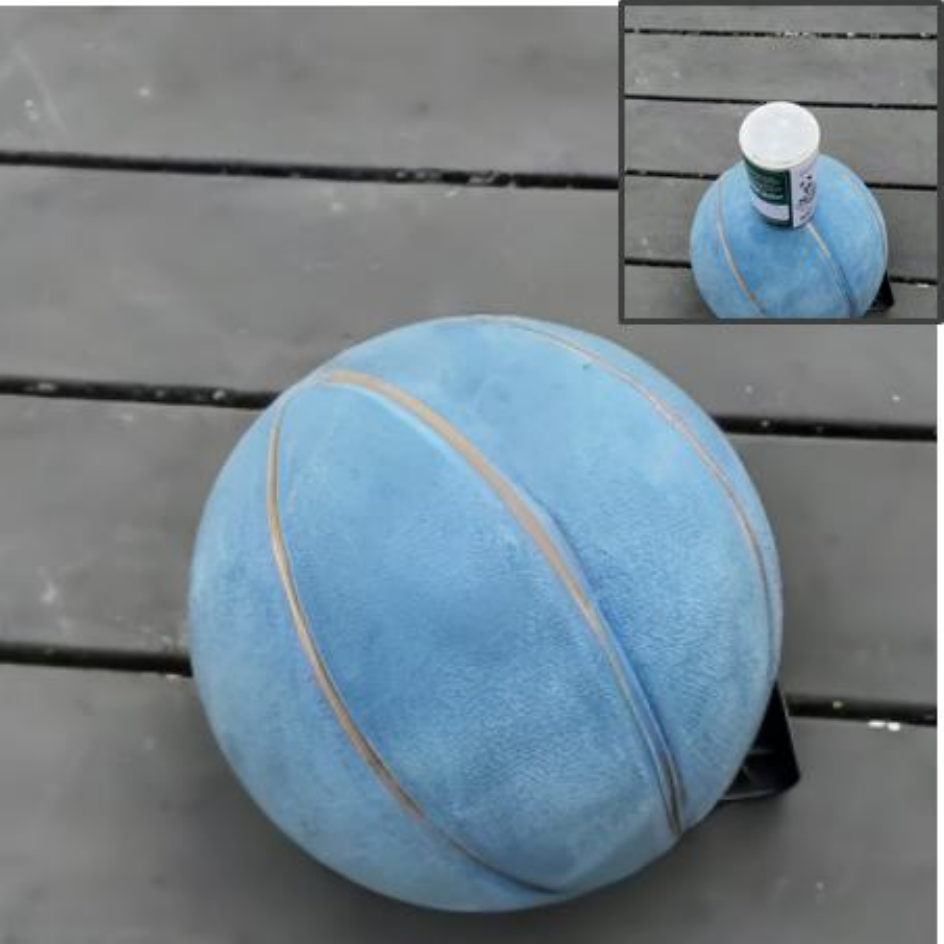} 
        \caption{Reference Image}
    \end{subfigure}
    \begin{subfigure}{0.15\textwidth}
        \centering
        \includegraphics[width=\linewidth]{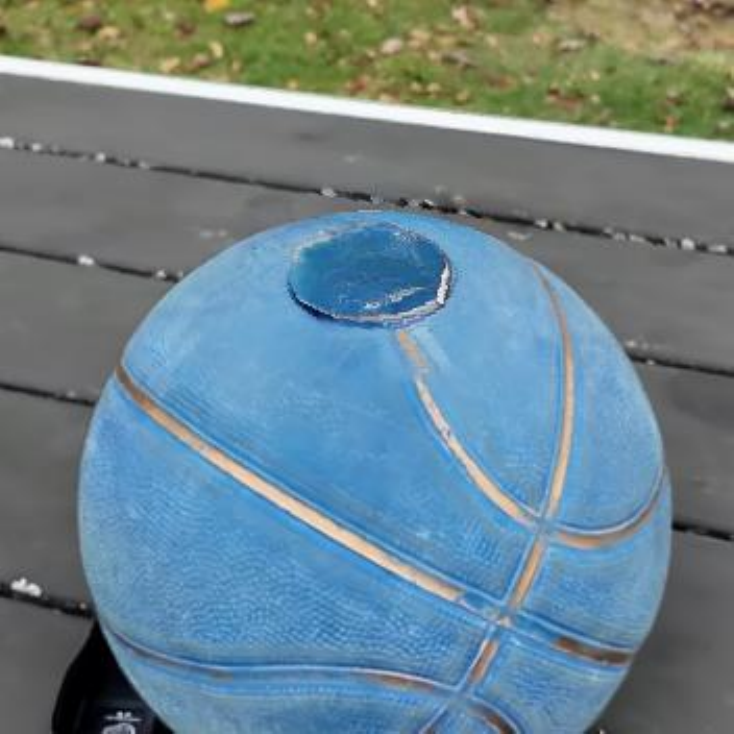} 
        \caption{W/o warping}
    \end{subfigure}
    \begin{subfigure}{0.15\textwidth}
        \centering
        \includegraphics[width=\linewidth]{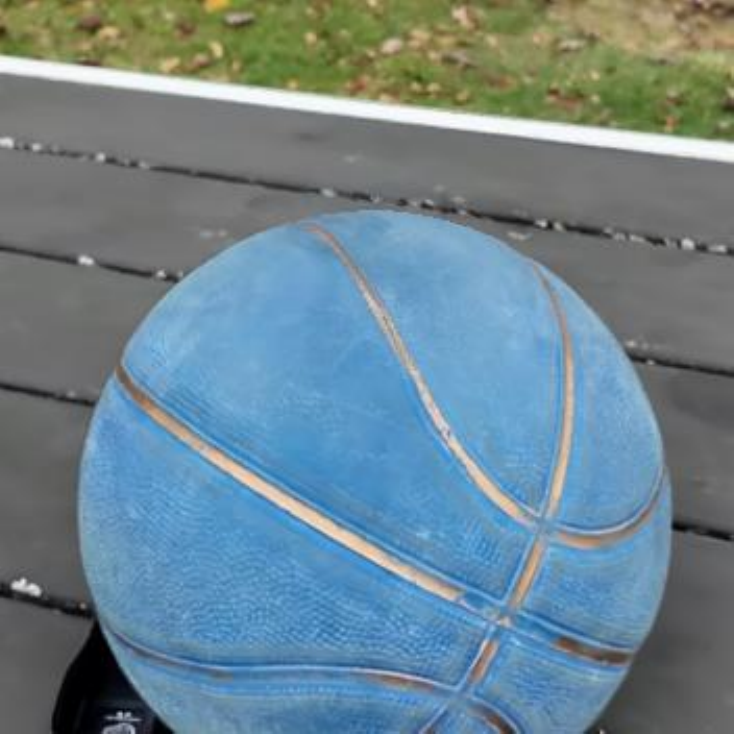} 
        \caption{W/ warping}
    \end{subfigure}
    \figvspace
    \caption{\textbf{Multi-view refinement with and without conditioning on the warped image.} (a) shows the inpainted reference with the originally captured image shown in the top right corner. (c) and (b) are the images from an alternate view generated with and without having the warped image as a condition, respectively.}
    \label{fig:ablate warping}
\end{figure}

We compare images generated with and without the warping process in Fig.~\ref{fig:ablate warping}. Despite the strength of multi-view architecture, direct inpainting fails to generate consistent output with the reference image in large view-change cases, which aligns with the observation in \cite{cao2024mvinpainter}. In contrast, our warping-refinement scheme provides a significant advantage, as the warped content offers strong guidance and regularizes the model to generate consistent results. We then use the generated 
images to fine-tune 3D scene. As shown in Tab.~\ref{tab:ablation}, the fine-tuned scene indeed shows inferior results to our complete method.

\begin{figure}[h]
    \centering    
    \begin{subfigure}{0.15\textwidth}
        \centering
        \includegraphics[width=\linewidth]{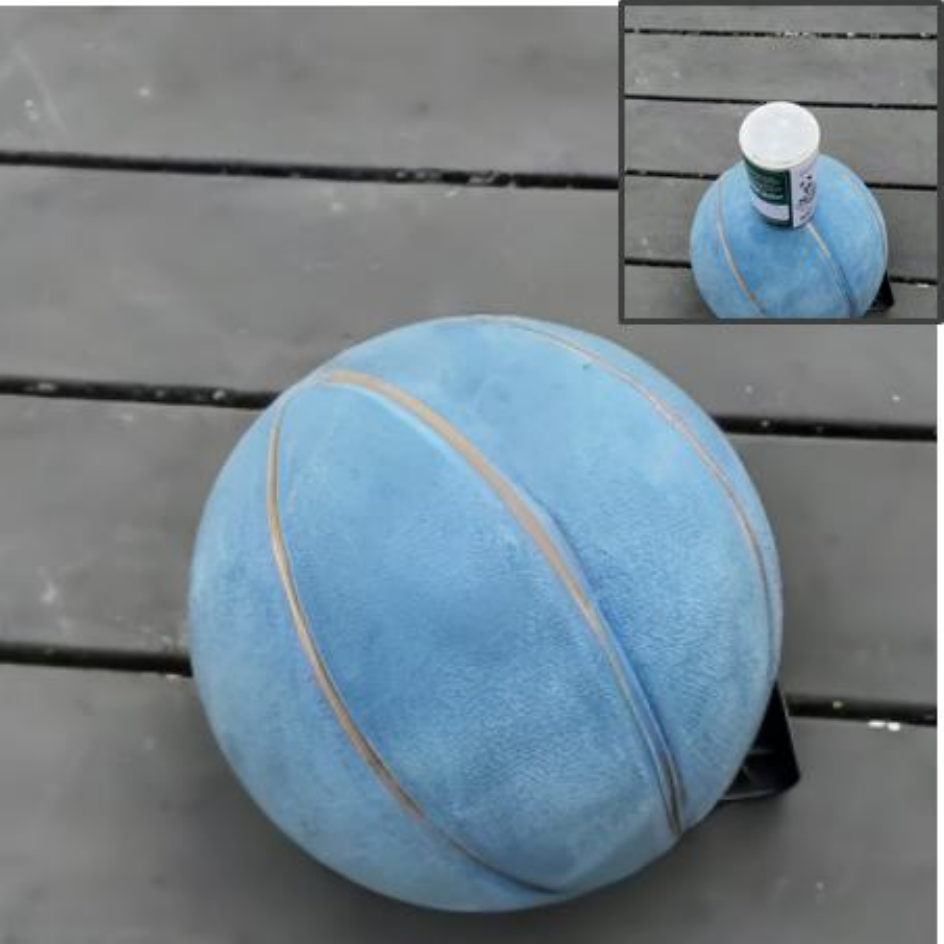} 
        \caption{Reference Image}
    \end{subfigure}
    \begin{subfigure}{0.15\textwidth}
        \centering
        \includegraphics[width=\linewidth]{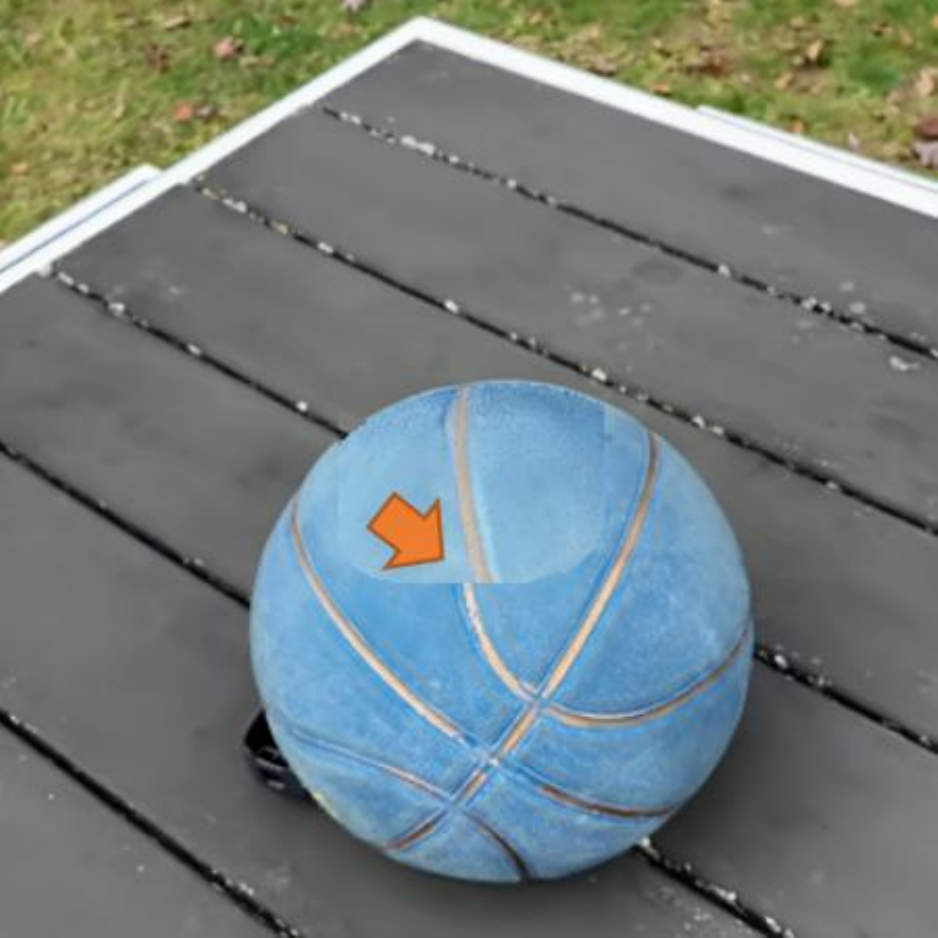} 
        \caption{W/o refine}
    \end{subfigure}
    \begin{subfigure}{0.15\textwidth}
        \centering
        \includegraphics[width=\linewidth]{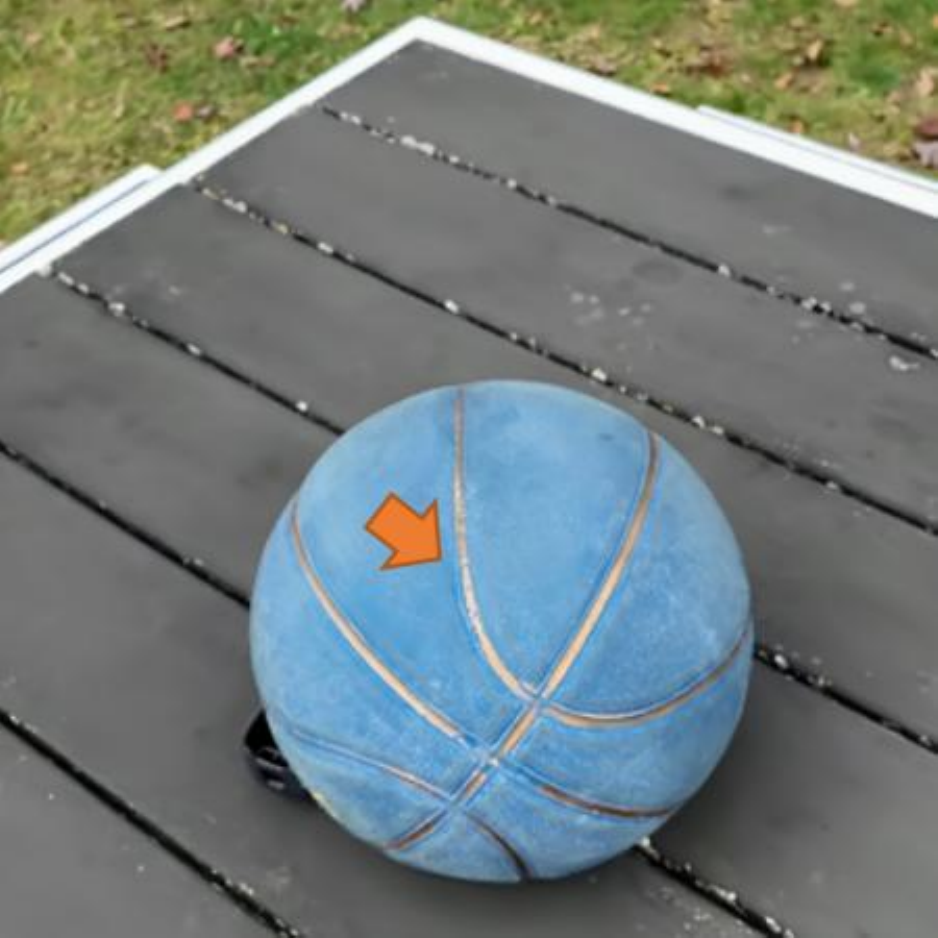} 
        \caption{Multi-view refine}
    \end{subfigure}
\\
    \begin{subfigure}{0.15\textwidth}
        \centering
        \includegraphics[width=\linewidth]{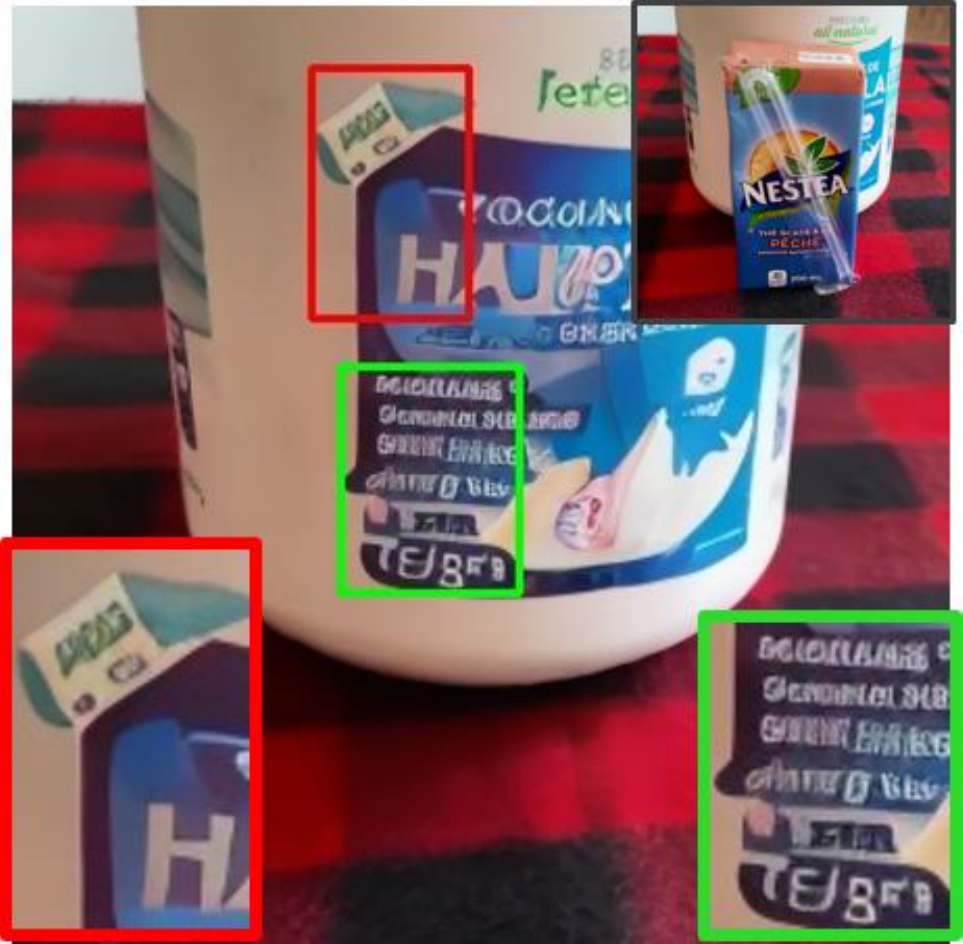} 
        \caption{Reference image}
    \end{subfigure}
    \begin{subfigure}{0.15\textwidth}
        \centering
        \includegraphics[width=\linewidth]{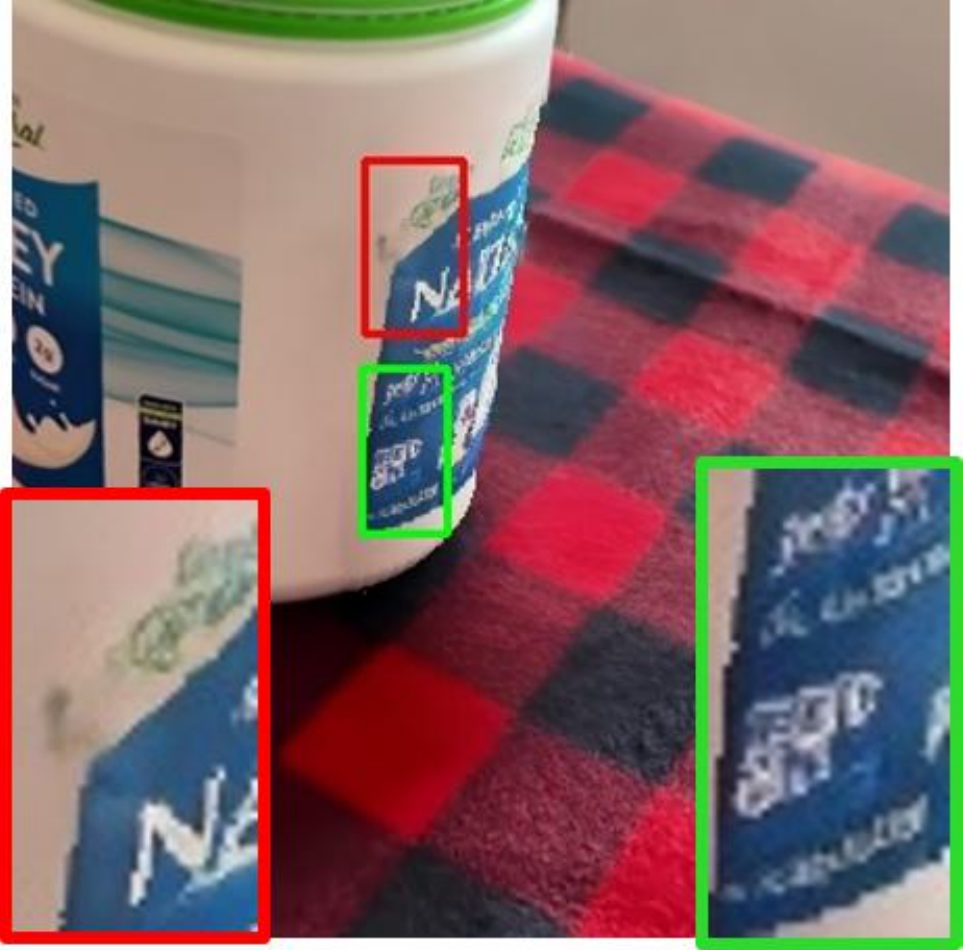} 
        \caption{Single-view refine}
    \end{subfigure}
    \begin{subfigure}{0.15\textwidth}
        \centering
        \includegraphics[width=\linewidth]{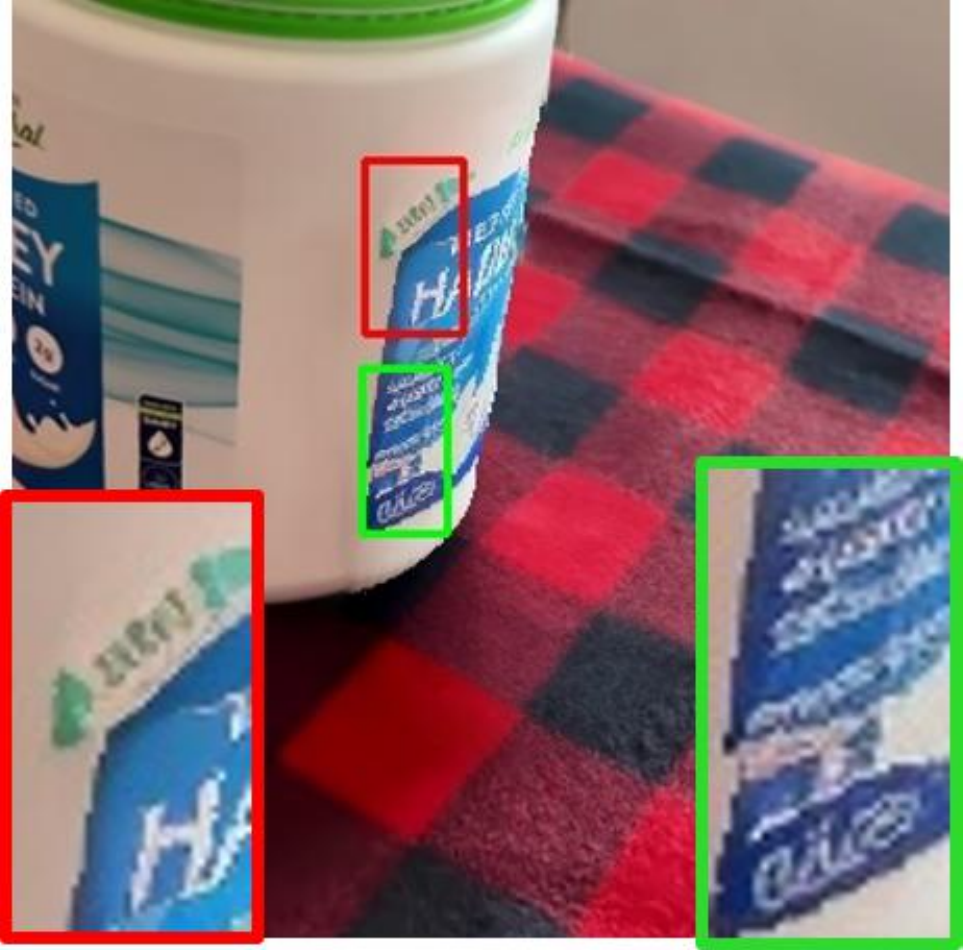} 
        \caption{Multi-view refine}
    \end{subfigure}
    \figvspace
    \caption{\textbf{Ablation on multi-view refinement.} (a) and (d) show the inpainted reference images with the originally captured images shown in the top right corner; (b) and (c) are the images from an alternate view generated without and with the refinement network, respectively; (e) and (f) are the images generated by the single-view and multi-view refinement model, respectively.}
    \vspace{-3mm}
    \label{fig:ablate refine}
\end{figure}

\noindent \textbf{Effectiveness of multi-view refinement model:}
We design the following two variants to illustrate the importance of the multi-view refinement model. a) w/o refinement: the refinement model is removed, and the warped images are directly used to fine-tune the pruned scene. b) single view: we replace the space-time attention with the original self-attention layer, performing refinement independently for each view without leveraging correlations across different views. When fine-tuning the single-view model for a given scene, only random image pairs are used.

Similarly, we compare images generated by these variants in Fig.~\ref{fig:ablate refine}. As seen in (a), (b), and (c), warping indeed introduces artifacts and texture mismatches, while the multi-view refinement model effectively addresses them. In (d), (e), and (f), the single-view refinement model fails to generate results consistent with the reference image, whereas the multi-view model achieves consistency. We further fine-tune the 3D scene with corresponding images. As illustrated in Tab.~\ref{tab:ablation}, our completed method achieves the best result.

\noindent \textbf{Necessity of test-time adaption:} 
Our pipeline requires a per-scene fine-tuned refinement model to generate images.
To demonstrate its necessity, we also train a generalized refinement model on the DI3DV dataset, a scene-level multi-view dataset, where we randomly selected two views as the reference and target, synthesized warping artifacts on one, and trained the model to restore. However, as shown in Tab.~\ref{tab:ablation}, its performance is inferior to our proposed method due to the limited size of the training set and domain gaps between the training set and real-world cases, such as differences in the severity of artifacts/mismatch.

\begin{figure}[h]
    \centering

    \begin{subfigure}{0.15\textwidth}
        \centering
        \includegraphics[width=\linewidth]{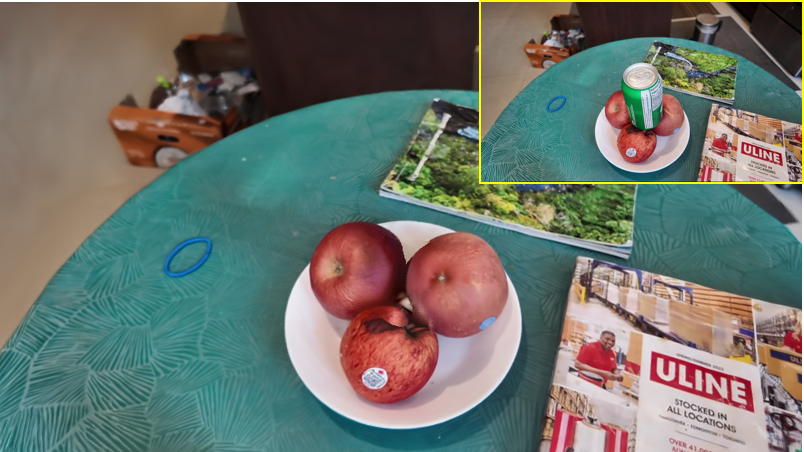} 
        \caption{Reference}
    \end{subfigure}
    \begin{subfigure}{0.15\textwidth}
        \centering
        \includegraphics[width=\linewidth]{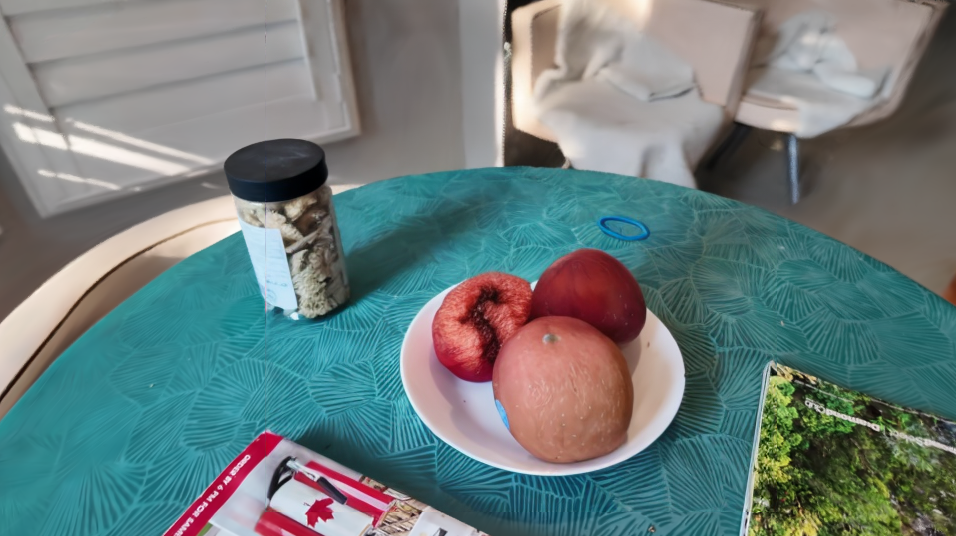} 
        \caption{1st View}
    \end{subfigure}
    \begin{subfigure}{0.15\textwidth}
        \centering
        \includegraphics[width=\linewidth]{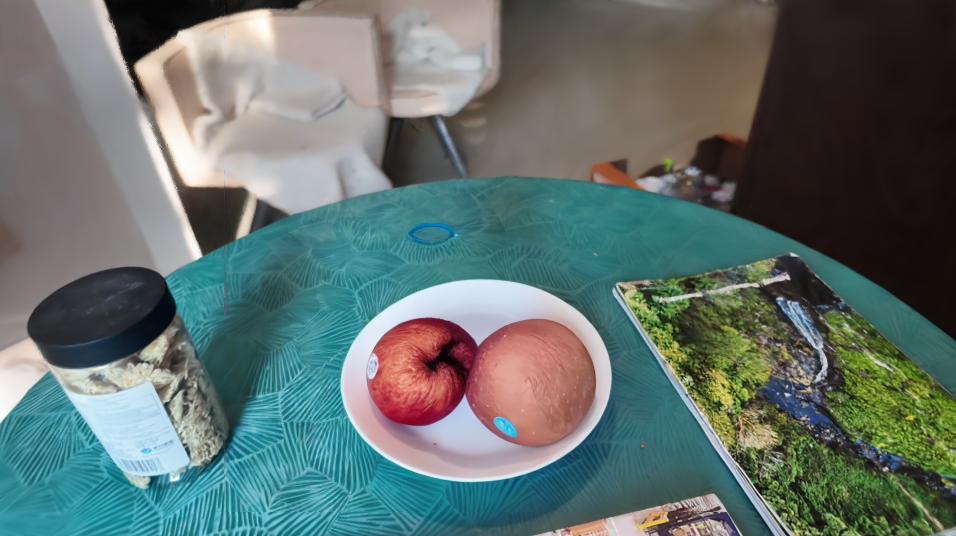} 
        \caption{2nd View}
    \end{subfigure}
    
    \figvspace
    \caption{Inpainting results by LeftRefill.}
    \label{fig:leftrefill}
    \vspace{-5mm}
\end{figure}

In addition, we use an existing inpainting method, LeftRefill \cite{cao2024leftrefill}, to directly inpaint images based on the reference image. However, as shown in Fig.\ref{fig:leftrefill}, LeftRefill can only ensures overall visual similarity but lacks pixel-level consistency.

\section{Conclusion and Limitation}
We present a novel 3D inpainting pipeline that achieves state-of-the-art results across both front-facing and unconstrained benchmarks, with particularly strong performance on the latter. We achieve this by utilizing geometry priors and a multi-view refinement network to generate multi-view consistent inpainted images. Additionally, a novel inpainting mask detection technique is developed to boost performance in unconstrained scenes. The effectiveness of our method is demonstrated through extensive experiments and ablation studies.

\noindent\textbf{Limitations:} The multi-view refinement network requires customization for each scene, which may take considerable time. Future advancements in parameter-efficient fine-tuning and the personalization of diffusion models may help to accelerate and further optimize our approach.
In addition, our framework relies on monocular depth estimation and image segmentation modules. The accuracy and robustness of these modules can affect the overall performance.

{
    \small
    \bibliographystyle{ieeenat_fullname}
    \bibliography{main}
}

\onecolumn
\begin{center}
    \textbf{\Large IMFine: 3D \underline{I}npainting via Geometry-guided \underline{M}ulti-view Re\underline{fine}ment}
    \vspace{0.5em} \\
    \textbf{\large - Supplementary Material -}
\end{center}
\vspace{-5pt}
\appendix
\setcounter{page}{1}

\section{Network Architecture}

The network input follows the same structure as Stable Diffusion inpainting, where an encoded warped image, a downsampled binary mask, and a random noise map are concatenated. Multiple views are processed in parallel by concatenating them along the batch dimension. Additionally, features from different views interact through our updated attention layers, referred to as space-time attention. These layers enable each view to attend to neighboring views and a specific reference view, ensuring view consistency.


\section{Fine-tuning Data Synthesis}
\begin{figure*}[t]
    \centering    
        \includegraphics[width=\linewidth]{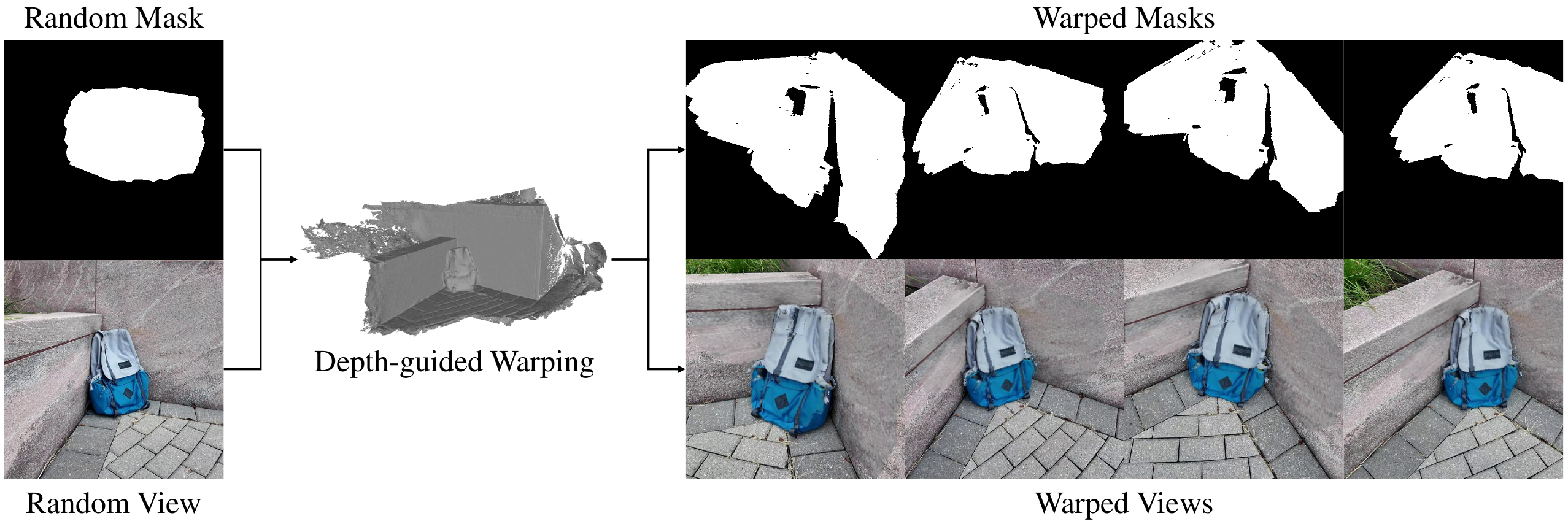} 
    \caption{Data synthesis based on depth-guided warping.}
    \label{fig:data_synthesis1}
\end{figure*}

In Fig.~\textcolor{red}{2}(b) of the main paper, we illustrate that the data used to fine-tune the multi-view refinement model is synthesized from the randomly selected view. Specifically, as illustrated in Fig.~\ref{fig:data_synthesis1}, we begin with the extracted 3D mesh from the unedited GS scene. An arbitrary view is chosen, and a random mask is generated around the target object. This view and the corresponding mask are then warped to other views, guided by the extracted mesh. 

\begin{figure*}[t]
    \centering    
        \includegraphics[width=0.72\linewidth]{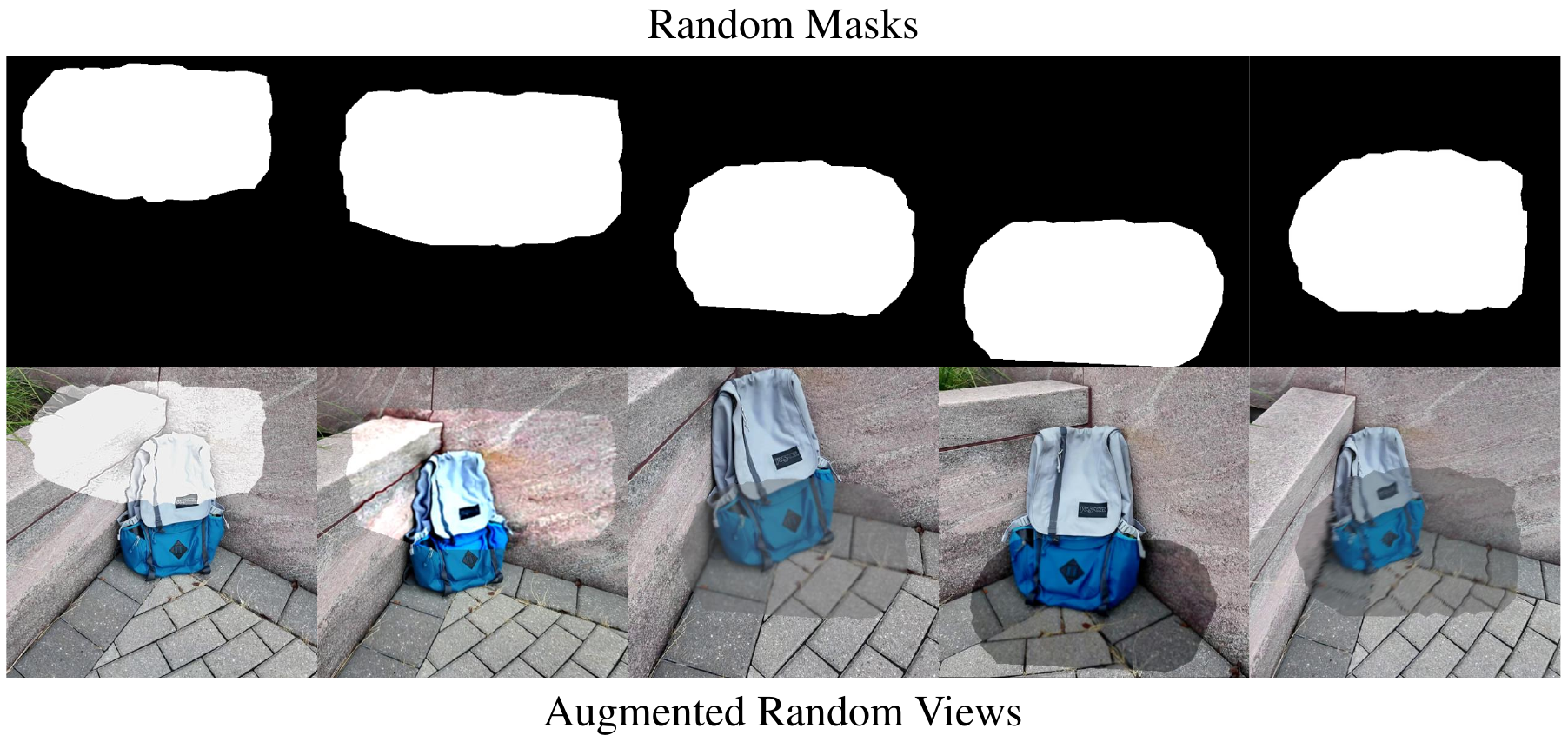} 
    \caption{Data synthesis based on image augmentation.}
    \label{fig:data_synthesis2}
\end{figure*}

To promote cross-view learning, we also generate an additional set of paired images by independently applying image augmentations to each view. As shown in Fig.~\ref{fig:data_synthesis2}, an irregularly shaped mask is randomly generated on a randomly selected view. Image-based augmentations—such as elastic transformations and color jittering, are then applied to the masked region to simulate warping artifacts.

\section{Dataset Description}
\begin{figure*}[t]
    \centering    
        \includegraphics[width=\linewidth]{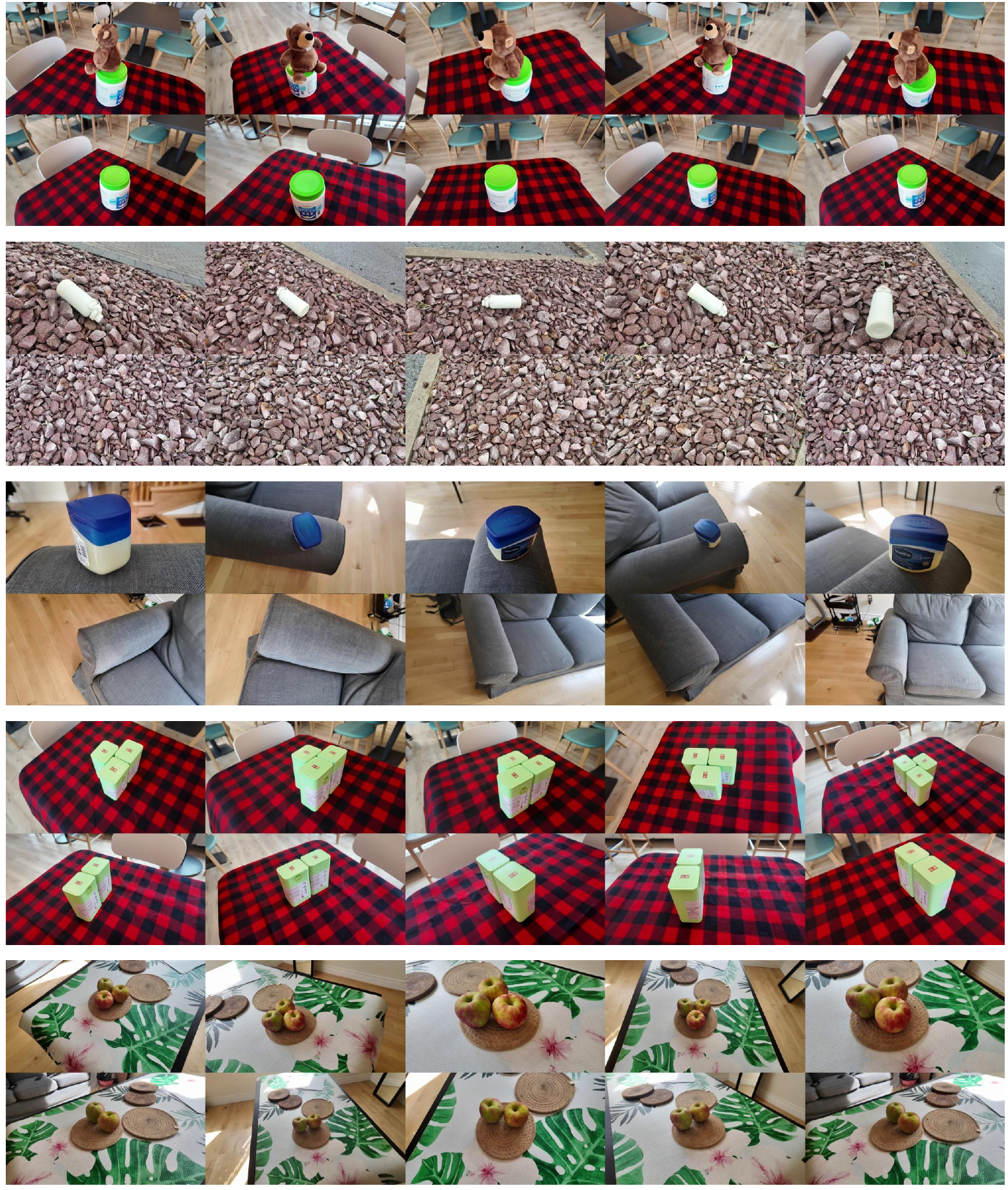} 
    \caption{Dataset visualization - 1.}
    \label{fig:dataset_all1}
\end{figure*}

\begin{figure*}[t]
    \centering    
        \includegraphics[width=\linewidth]{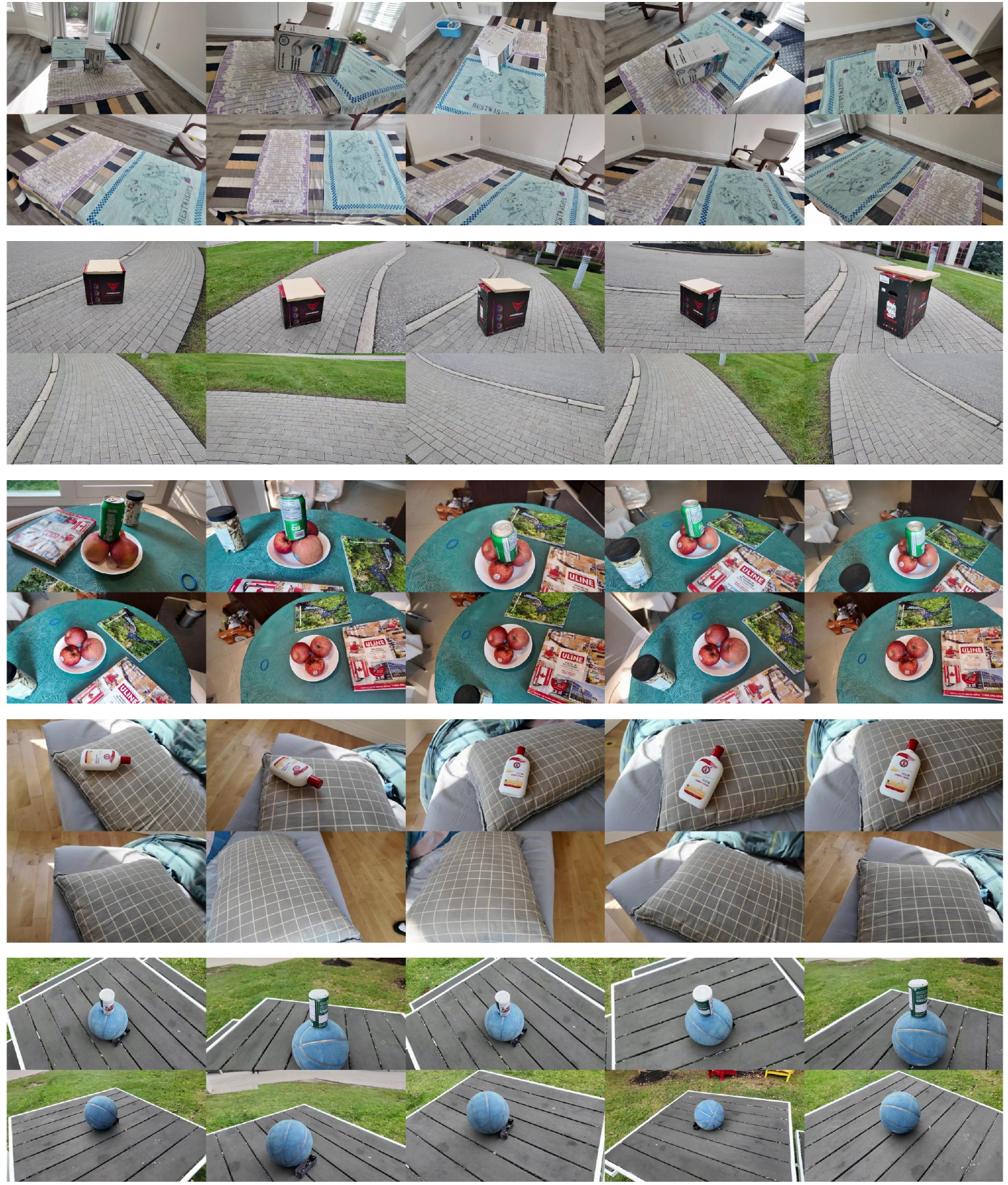} 
    \caption{Dataset visualization - 2.}
    \label{fig:dataset_all2}
\end{figure*}

\begin{figure*}[t]
    \centering    
        \includegraphics[width=\linewidth]{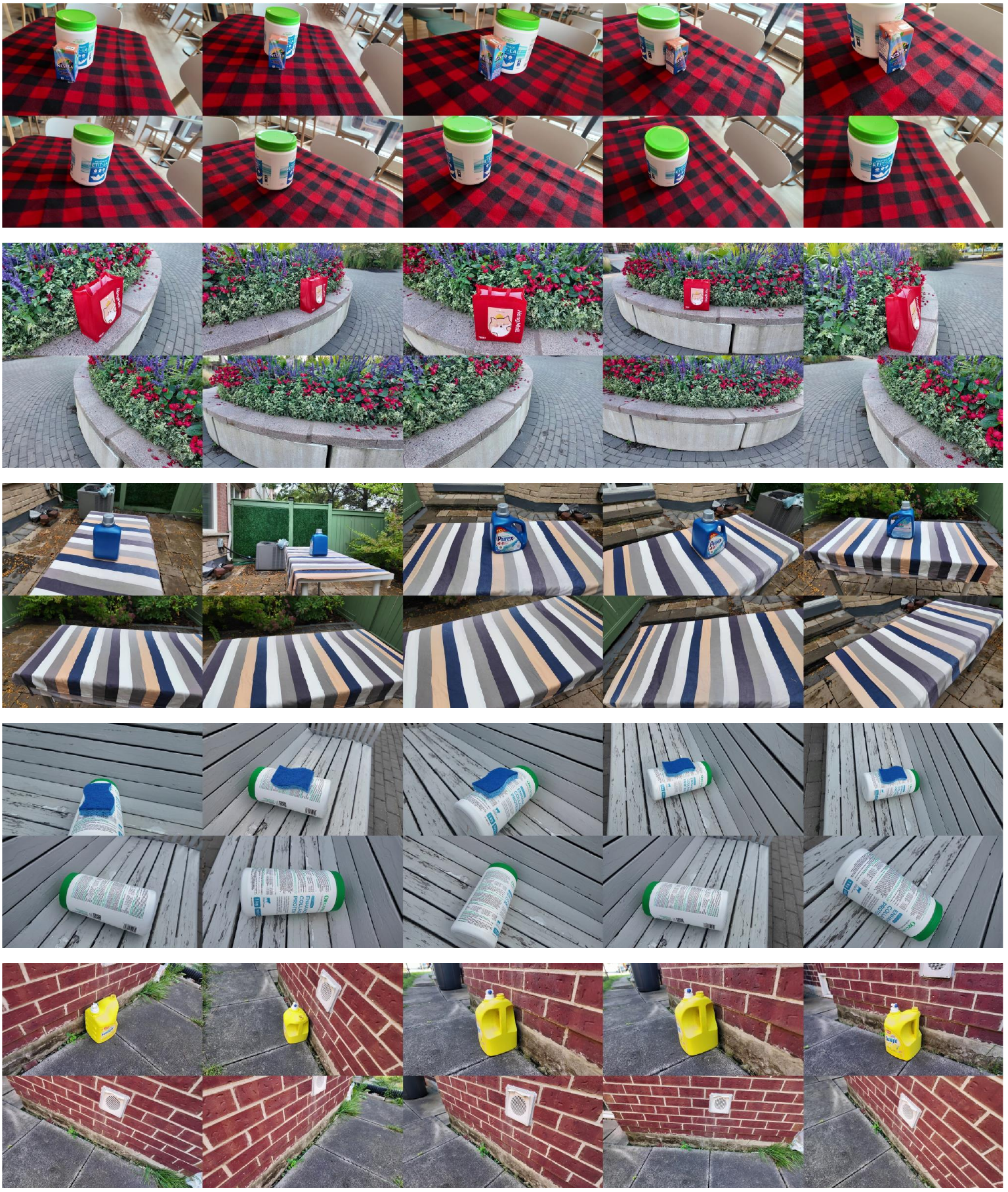} 
    \caption{Dataset visualization - 3.}
    \label{fig:dataset_all3}
\end{figure*}

\begin{figure*}[t]
    \centering    
        \includegraphics[width=\linewidth]{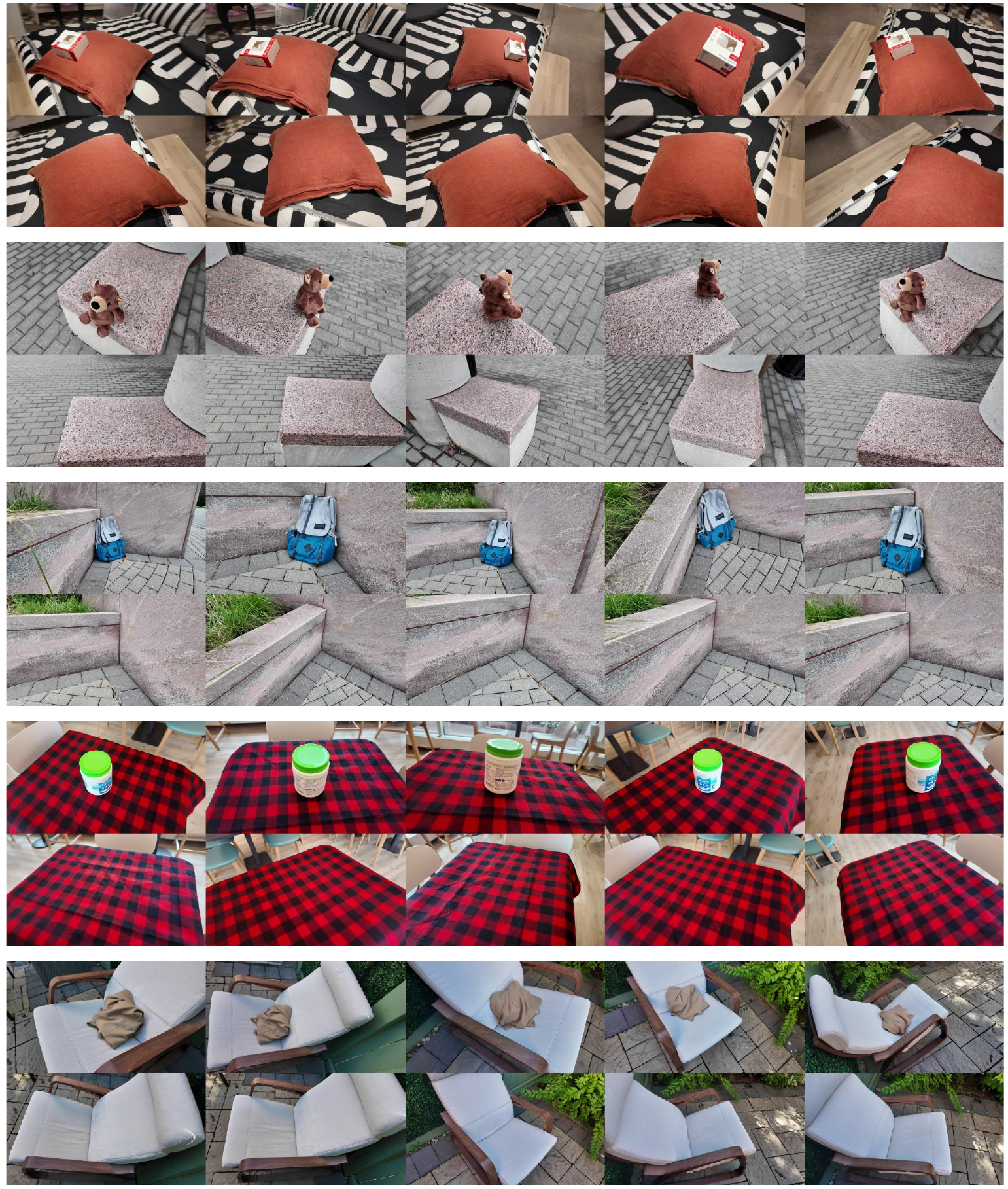} 
    \caption{Dataset visualization - 4.}
    \label{fig:dataset_all4}
\end{figure*}

The proposed dataset comprises 20 scenes with diverse characteristics. Specifically, there are 2, 4, 4, and 10 scenes where the camera trajectory spans 90°, 120°, 180°, and 360°, respectively. Additionally, the dataset includes 4, 5, 5, and 6 scenes featuring single-plane, curved-plane, multi-plane, and irregular interfaces. Each scene consists of 175 images, along with corresponding camera poses and object masks. The first 50 images are used for evaluation purposes with the object removed, while the remaining 125 images containing the object are utilized to reconstruct the original Gaussian Splatting scene (GS). More data visualization is shown in Fig.~\ref{fig:dataset_all1} to~\ref{fig:dataset_all4}.

Object masks for the training views are generated using SAM2, with human-provided masks for the first frame. For the testing views, object masks are derived through a three-step process: (1) reconstructing a GS scene using the training images, (2) segmenting object Gaussians based on the training object masks, and (3) rendering the object masks for the test views.

\section{More Qualitative Evaluations}
We provide additional qualitative evaluations and failure cases in the attached video. Our method has limitations when the original 3D scene is poorly reconstructed. Under such conditions, significantly anisotropic or oversized Gaussian primitives may become exposed in the pruned scene, negatively impacting subsequent processing.



\end{document}